\documentclass{article}

 \usepackage[preprint]{neurips_2026}

\usepackage[utf8]{inputenc} %
\usepackage[T1]{fontenc}    %
\usepackage{hyperref}       %
\usepackage{url}            %
\usepackage{booktabs}       %
\usepackage{amsfonts}       %
\usepackage{nicefrac}       %
\usepackage{microtype}      %
\usepackage{xcolor}         %
\usepackage{graphicx}
\usepackage{amsmath}
\usepackage{amssymb}
\usepackage{booktabs}
\usepackage{adjustbox}
\usepackage{multirow}
\usepackage[table]{xcolor}
\usepackage{array}
\usepackage{makecell} %
\usepackage{wrapfig}
\usepackage{fontawesome5}
\usepackage{svg}

\definecolor{CiteBlue}{HTML}{1F4E79}
\definecolor{FACSPlum}{HTML}{1B5E20}
\definecolor{TaskBurgundy}{HTML}{7A1F2B}

\definecolor{lightblue}{RGB}{222,235,247}
\definecolor{lightpurple}{RGB}{233,222,247}
\definecolor{lightgreen}{RGB}{222,247,233}
\definecolor{lightorange}{RGB}{252,236,205}
\definecolor{lightgray}{RGB}{240,240,240}
\usepackage{array}
\usepackage[most]{tcolorbox}
\usepackage{listings}
\usepackage{titletoc}
\usepackage{graphicx}
\usepackage{subcaption}
\usepackage{caption}
\usepackage{dblfloatfix}
\usepackage{tabularx}
\usepackage{cleveref}
\usepackage{longtable}
\usepackage{placeins}

\title{K9-Bench: Evaluating Multimodal LLMs on Canine-Centric Videos}

\tcbset{
  colback=blue!4!white,
  colframe=black!60,
  boxrule=0.3pt,
  arc=2pt,
  left=4pt,
  right=4pt,
  top=4pt,
  bottom=4pt,
  coltext=black,        %
}

\lstdefinestyle{promptstyle}{
    basicstyle=\normalfont\small,  %
    columns=fullflexible,
    breaklines=true,
    showstringspaces=false,
    keepspaces=true,
    keywordstyle=\color{black},
    commentstyle=\color{black},
    stringstyle=\color{black},
    identifierstyle=\color{black},
    moredelim=[is][\color{black}]{\%}{\%},
}

\lstdefinestyle{smallpromptstyle}{
    basicstyle=\normalfont\scriptsize,  %
    columns=fullflexible,
    breaklines=true,
    showstringspaces=false,
    keepspaces=true,
    keywordstyle=\color{black},
    commentstyle=\color{black},
    stringstyle=\color{black},
    identifierstyle=\color{black},
    moredelim=[is][\color{black}]{\%}{\%},
}

\vspace{-10mm}
\author{%
  Khush Attarde$^{*1}$ \quad
  Yusuf Ali$^{*2}$ \quad
  Megha Thukral$^{\dagger2}$ \quad
  Divye Bhutani$^{1}$ \\
  \textbf{Thomas Ploetz}$^{2}$ \quad
  \textbf{Zsolt Kira}$^{2}$ \vspace{1.5mm} \\ 
  $^{1}$Ogmen Robotics Inc., Massachusetts, USA \\
  $^{2}$Georgia Institute of Technology, Atlanta, USA \vspace{0.8mm}
  \\
  \texttt{\{khush.attarde,divye\}@ogmenrobotics.com} \\
  \texttt{\{yali30,mthukral3,thomas.ploetz,zkira\}@gatech.edu}
}

\begin{document}

\makeatletter
\renewcommand{\footnoterule}{%
  \kern -3pt
  \hrule \@width 0.35\textwidth
  \kern 2.6pt
}
\makeatother

\begingroup
\renewcommand{\thefootnote}{}
\footnotetext{\textsuperscript{*}Co-first authors. \textsuperscript{$\dagger$}Core Contributor.}
\endgroup

\maketitle

\definecolor{cbf}{HTML}{7ED957}
\definecolor{cspu}{HTML}{B395FF}
\definecolor{csrd}{HTML}{FF70A6}

\vspace{-3mm}
\begin{abstract}

\vspace{-2mm}
Multimodal Large Language Models (MLLMs) have demonstrated remarkable zero-shot capabilities across diverse inputs such as images, video, audio, and text. 
A crucial, yet underexplored, application of these models lies in understanding and modeling animal-centric scenarios, especially domesticated animals. 
As animals are integral to millions of households, benchmarking next-generation AI models on pet-focused tasks, ranging from recognizing distress signals in pets to enabling responsive robotic companions, is essential for building AI systems that can live and work alongside us.
We introduce \texttt{K9-Bench}, a novel benchmark focused on real-world videos of domestic dogs, specifically targeting canine action and interaction understanding via $\approx$5000 question-answer pairs
across $907$ videos spanning $5$ distinct task categories that test long-form, canine-centric multimodal reasoning in MLLMs. 
To create this dataset, we propose a scalable, VLM/LLM-powered data generation pipeline that automatically mines canine-centric videos from open web sources and curates QA pairs requiring fine-grained, multi-hop reasoning over canine actions and temporally extended interaction sequences.
We further propose bias mitigation strategies designed to eliminate biases introduced by VLMs during dataset curation.
Through extensive experimentation, we find that frontier MLLMs exhibit limited zero-shot performance on canine-centric tasks: although state-of-the-art closed-source models outperform open-source counterparts, they still struggle with compositional reasoning over subtle posture and interaction cues spread over long horizons.
We further observe that generic chain-of-thought prompting provides only modest performance for such long-horizon reasoning.
We also conduct human evaluations and checks on a subset to validate the overall dataset quality.
Beyond a novel dataset for canine activity analysis, \texttt{K9-Bench} provides a general-purpose dataset construction pipeline that can be adapted to other \emph{low-data domains} for quantitative analysis.
Our project website, including links to the dataset and evaluation suite, is available \href{https://ogmenrobotics.github.io/K9Bench/}{here}.
\end{abstract}

\vspace{-3mm}
\section{Introduction}

\begin{figure}[h]
    \centering
    \includegraphics[width=\columnwidth]{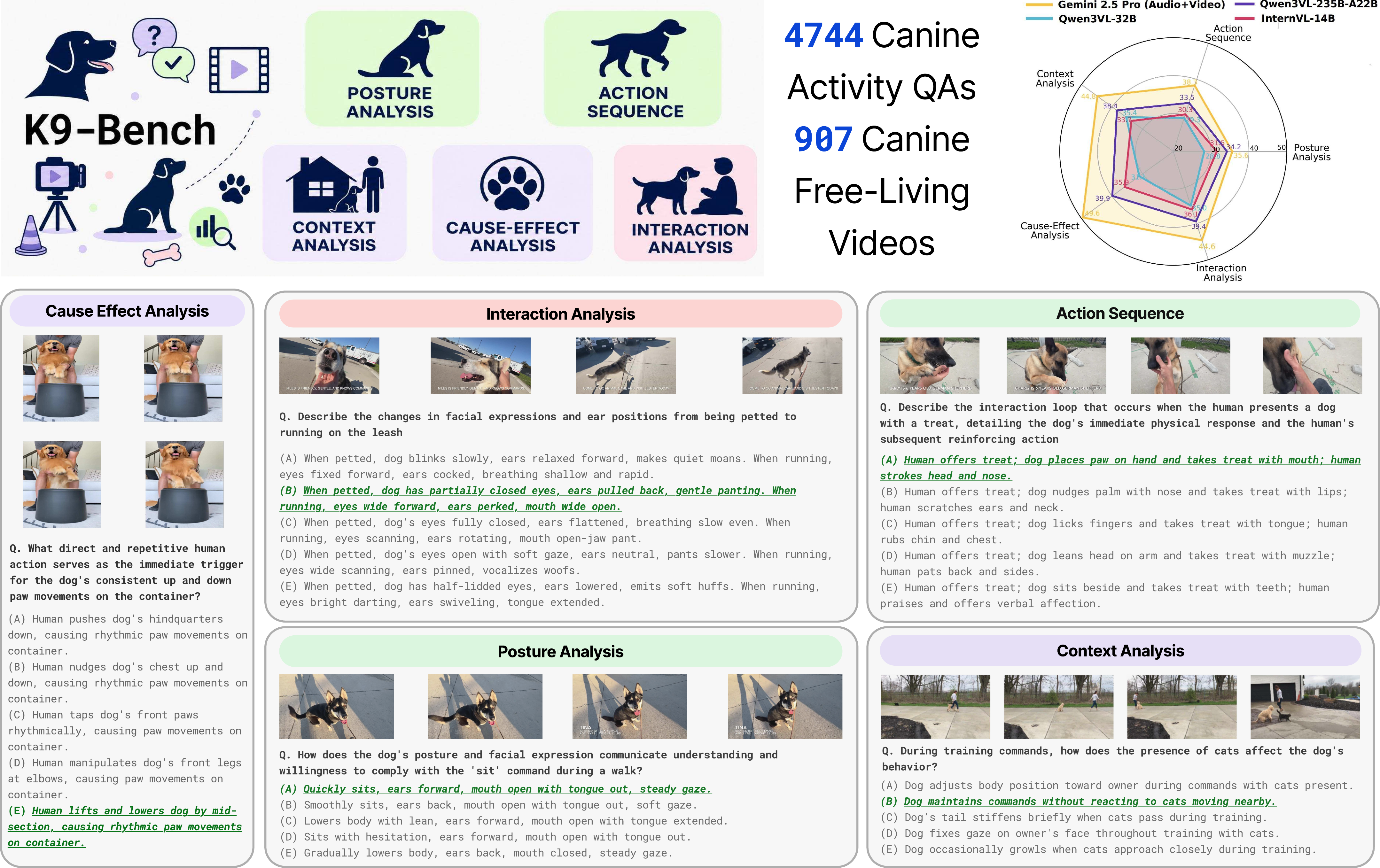}
    \caption{We introduce \texttt{\textbf{K9-Bench}}: a novel canine activity and interaction-focused benchmark that evaluates frontier multimodal LLMs on canine-focused video question-answering tasks.}
    \label{fig:main_fig}
    \vspace{-7.5mm}
\end{figure}

\vspace{-3  mm}
The growing integration of intelligent systems into everyday human environments motivates the development of AI agents that can perceive, interpret, and reason over rich multimodal streams, not only in human-centric scenarios, but also in settings that involve the animals that share our homes, streets, and public spaces. 
Dogs, in particular, are one of the most prevalent and socially integrated non-human species: an estimated $71\%$ of U.S.\ households have pets, and roughly $68$ million households include at least one dog~\cite{american2025american}. Building AI systems that can robustly understand canine activities and interactions is therefore central to applications ranging from pet safety and welfare monitoring to assistive home robotics and interactive companions.

With rapid progress in MLLM capabilities, a growing set of benchmarks has been developed to evaluate long-form video understanding, temporal reasoning, and multimodal grounding~\cite{xiao2021next,li2024mvbench}. 
Recent models now support unified video–audio–text reasoning at scale~\cite{team2024gemini,xu2025qwen3}. 
However, these evaluations remain predominantly human-centric, with tasks and data centered on human actions, cues, and social conventions. 
As a result, they offer limited insight into whether current models can generalize beyond familiar human contexts.
Animal-centric scenarios introduce different motion patterns, interaction cues, and sources of ambiguity. 
Prior work has compiled species-focused datasets and resources~\cite{jing2024animal,chen2023mammalnet}, and early attempts have begun integrating vision–language models into non-human domains~\cite{jing2025animal}. 
Yet these efforts typically involve narrow tasks or \emph{manual supervision}, leaving open how well modern MLLMs handle everyday human–animal interactions in realistic, unconstrained environments.

\vspace{-0.1mm}
In this work, we focus on \emph{canine-centric video question answering} with frontier MLLMs.
We introduce \texttt{K9-Bench}, a benchmark built from $907$ real-world videos of domestic dogs that targets action and interaction understanding across long temporal video sequences.
Our goal is to move beyond short, isolated clips or single-label classification toward fine-grained, multi-hop reasoning about how dogs move, react, and interact with humans and other dogs over time.
We generate questions and answers across $5$ canine-centric task categories, created with inputs from an expert canine-trainer, covering scenarios such as \textit{posture}, \textit{multi-action composition}, and \textit{causal reasoning} (see \Cref{fig:main_fig}).
To scale this setting beyond what is feasible with manual annotation, we propose a VLM/LLM-powered pipeline that automatically filters canine-centric videos mined from open web sources and constructs semantically rich question–answer pairs while explicitly addressing issues of semantic relevance, textual shortcuts, and model bias.

\vspace{-0.1mm}
\textbf{Contributions.}
Our work makes the following key contributions:
\vspace{-3mm}
\begin{itemize} \setlength{\itemsep}{0.08em}
    \item We formulate canine-centric video question answering as a \emph{challenging testbed for multimodal reasoning} in shared human–animal environments. \vspace{-1mm}
    \item We introduce \texttt{K9-Bench}, a new benchmark comprising $\approx5000$ video–QA pairs across $5$ tasks that probe long-horizon, multimodal understanding of everyday canine activities and interactions. \vspace{-1mm}
    \item We present a scalable dataset generation pipeline that leverages VLMs and LLMs to automatically curate videos and generate, filter, and refine question–answer pairs with a focus on semantic fidelity, reduced textual shortcuts, and bias mitigation. 
\end{itemize}

\section{Related Work}

\vspace{-3mm}
\paragraph{MLLMs for Video Understanding.} 
Multimodal large language models (MLLMs) have rapidly advanced in their general ability to process long videos \cite{team2024gemini,bai2025qwen2,zhang2024llavanextvideo,wang2025internvl3}.
Recent work in the computer vision literature has proposed a wide variety of domains to stress-test video understanding capabilities \cite{xiao2021next,mangalam2023egoschema,li2024mvbench,yang2025thinking}.
Video-MME \cite{fu2024video} curates videos across multiple domains with synchronized audio and subtitle streams to assess temporal and multimodal reasoning.
LongVideoBench \cite{wu2024longvideobench} focuses on assessing multi-frame referential reasoning capabilities over long videos.
Curating such benchmarks typically involves a significant amount of human effort which limits scalability \cite{chandrasegaran2024hourvideo,zhou2025mlvu,majumdar2024openeqa}, data diversity and restrictions to templated guidelines.
Few recent works have proposed leveraging LLM/VLM capabilities to build scalable dataset generation pipelines.
CinePile \cite{rawal2024cinepile} uses an adversarial LLM-based refinement routine to automatically generated question-answer pairs to maintain task difficulty and prevent shortcuts in solvability.
VideoEspresso \cite{han2025videoespresso} proposes an automated pipeline for QA generation with associated chain-of-thought-style explanations. 
VideoMarathon \cite{lin2025unleashing} generates a synthetic long video understanding dataset using a hierarchcial captioning strategy.
These works propose synthetic data generation pipelines using videos sourced from prior large-scale video understanding datasets which focus primarily on human-centric activities.
In this work, we instead propose a pipeline that enables automated QA generation of canine-centric videos with an explicit focus on maintaining the semantic relevance of generated QA pairs and mitigating inherent LLM bias.

\textbf{Animal-Centric Datasets and Benchmarking.} 
A lot of recent work has focused on developing large-scale datasets and novel methods for animal activity and behavior analysis.
AnimalBench \cite{jing2024animal} aggregates multiple existing animal-centric datasets and proposes an automated pipeline for video question-answer generation,
MammalNet \cite{chen2023mammalnet} developed a large-scale, human-annotated video dataset with labeled fundamental behaviors across multiple species.
MMAlps \cite{gabeff2025mammalps} collects a wildlife dataset comprising $5$ species and provides manual annotations with a two-level behavior hierarchy.
Animal Kingdom \cite{ng2022animal} curates a video dataset with manual annotation of atomic actions across $850$ animal species.
In contrast to the aforementioned, there are works that focus on specific animal species such as pigs \cite{liu2020computer}, horses \cite{mathis2021pretraining}, baboons \cite{duporge2025baboonland} and dogs \cite{khosla2011novel}.
But these works do not constitue long-form videos which can be leveraged to evaluate the multimodal reasoning capabilities of frontier MLLMs on these animal-centric sequences.
In this work, we build a novel video question-answering dataset that specifically focuses on everyday canine activities and interactions.

\textbf{MLLMs for Animal-Centric Analysis.} 
Recent work has begun using LLMs and VLMs in aiding automated behavior descriptions and monitoring for diverse species.
BehaveAgent \cite{aljovic2025autonomous} proposes an agentic workflow in which multiple foundation models are orchestrated to track and segment individual movements leading to behavior descriptions.
MouseGPT \cite{xu2025mousegpt} trains a VLM for mouse behavior analysis on data that was curated using an LLM as rating model to filter out high quality training instances.
Animal-CLIP \cite{jing2025animal} uses LLM to generate diverse prompts whose text embeddings are used in a contrastive learning routine to train a video-action recognition model.
VideoPrism \cite{sun2024video} finetunes a large, pretrained VLM on behavior classification datasets and outperforms specialist, in-domain baselines.
In this work, we leverage MLLMs to generate the initial set of seed question-answer pairs focused on canine activities.
We propose additional steps to refine these generated QA pairs to mitigate bias and textual shortcuts in solving the QAs.

\vspace{-5mm}
\section{\texttt{K9-Bench}} \label{sec:k9_bench_overview}

\begin{figure*}[t]
\centering

\begin{subfigure}{\textwidth}
    \centering
    \includegraphics[width=\textwidth,
    trim={30mm 40mm 25mm 40mm}, clip]{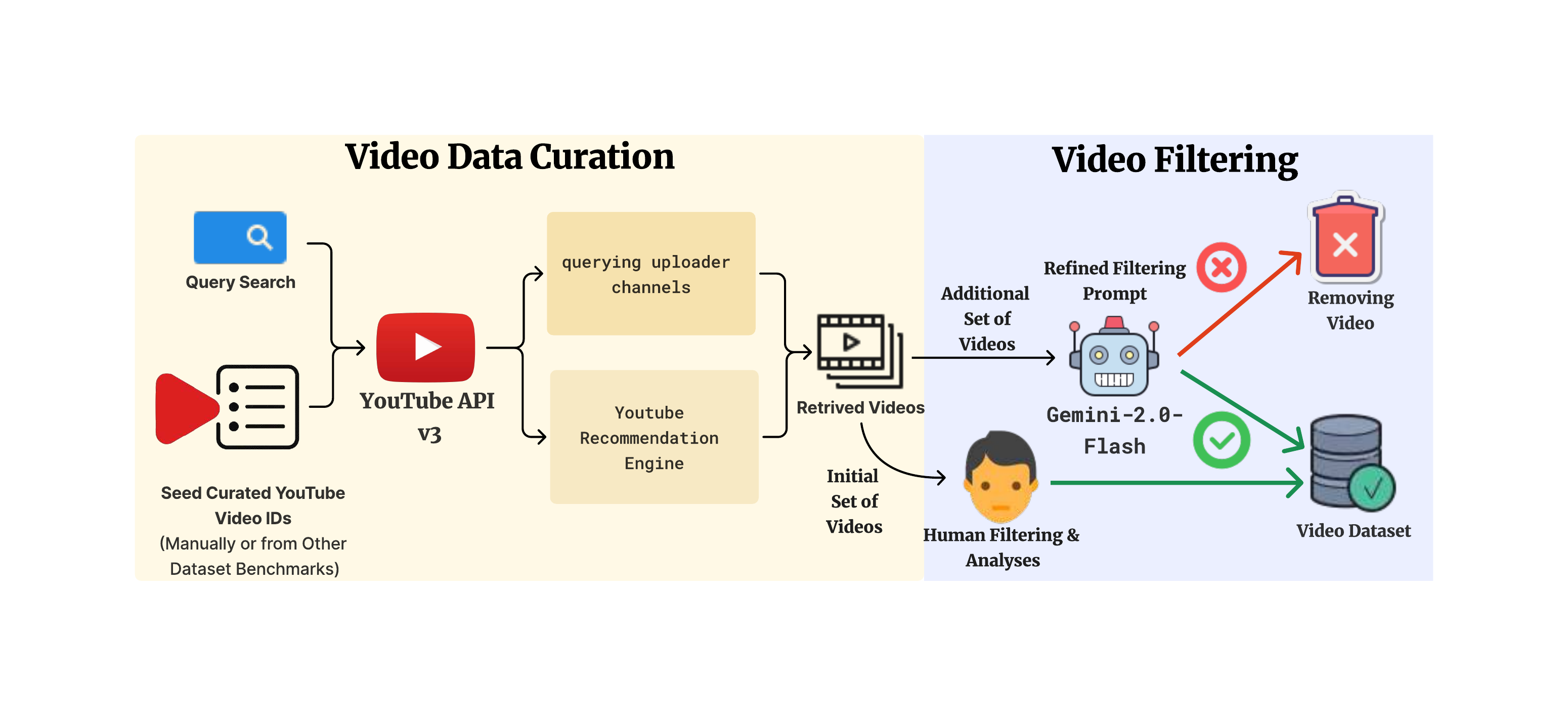}
    \subcaption{}
    \label{fig:vid_curation_pipeline}
\end{subfigure}

\vspace{-2mm}

\begin{subfigure}{\textwidth}
    \centering
    \includegraphics[width=\textwidth,
    trim={30mm 40mm 25mm 40mm}, clip]{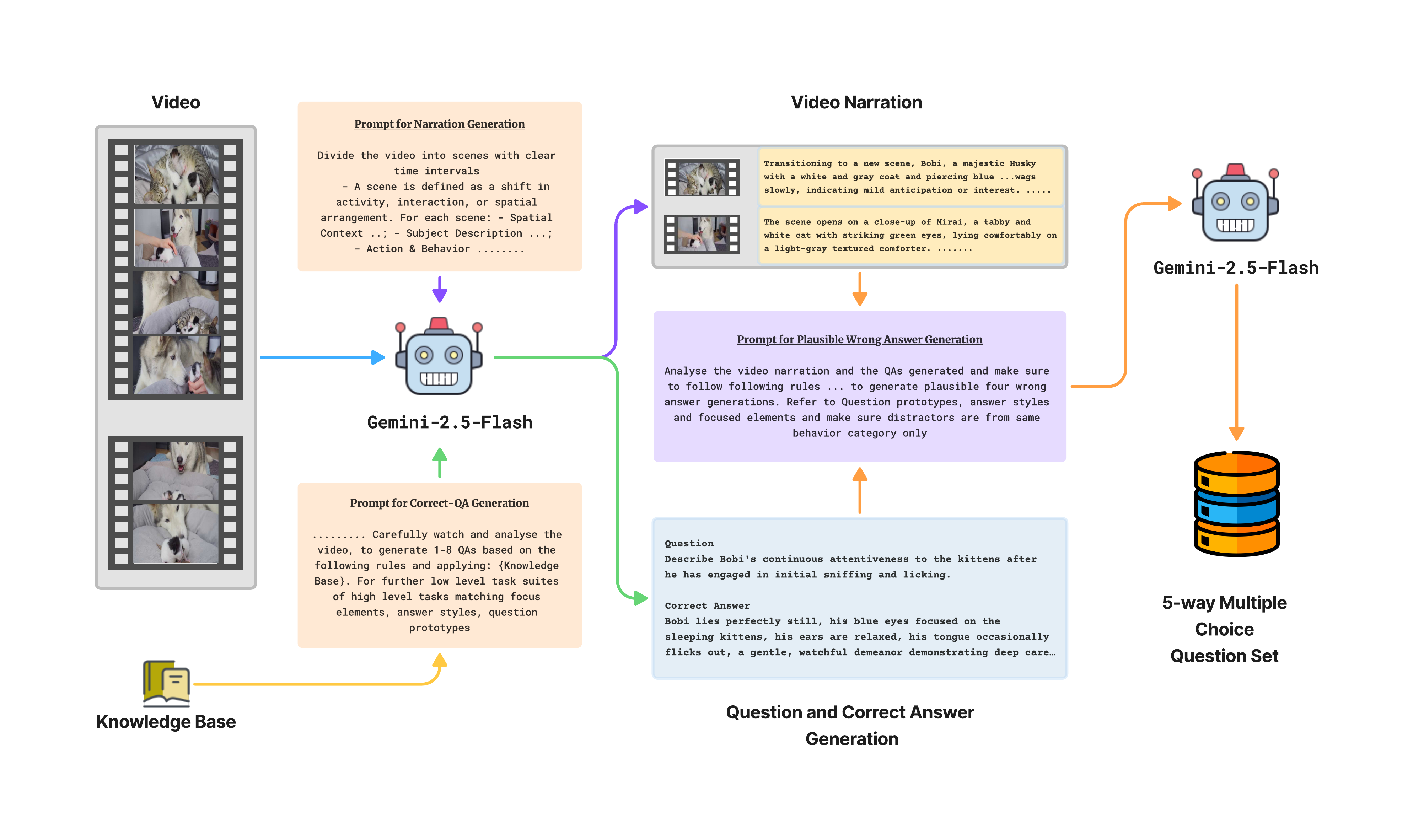}
    \subcaption{}
    \label{fig:qa_generation}
\end{subfigure}

\vspace{-2mm}

\begin{subfigure}{\textwidth}
    \centering
    \includegraphics[width=\textwidth,
    trim={30mm 40mm 25mm 40mm}, clip]{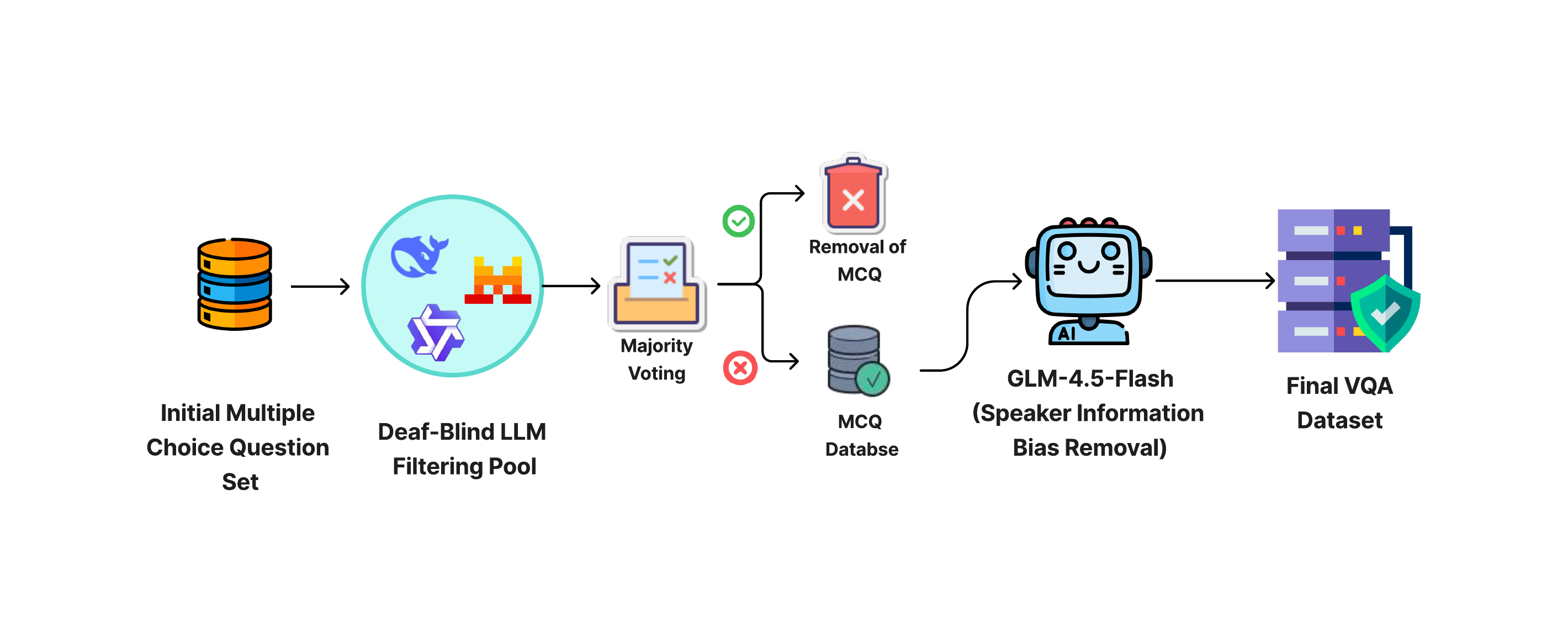}
    
    \subcaption{}
    \label{fig:bias_mitigation_pipeline}
\end{subfigure}

\vspace{-2mm}

\caption{\textbf{(a).} Video curation and filtering pipeline for \texttt{K9-Bench}. The initial and additional video sets correspond to the pre- and post-refined filtering stages. \textbf{(b)} 5-way multiple-choice question generation pipeline leveraging the vision-language model \texttt{Gemini2.5-Flash}. \textbf{(c)} Bias mitigation pipeline that filters Q\&A pairs solvable from text alone and removes speaker-identifying information.}
\label{fig:pipelines_combined}
\vspace{-5.5mm}

\end{figure*}

\vspace{-3mm}
We introduce \texttt{K9-Bench}, a benchmark designed to systematically evaluate the multi-modal understanding of fine-grained canine actions and interactions from real-world videos (see \Cref{fig:main_fig}).
The video QA dataset contains $4744$  five-way multiple-choice QA pairs derived from a curated subset of $907$ videos.
These videos come from a larger pool i.e. $1737$ videos, that undergo strict quality filtering via a \textbf{human-in-the-loop verification} process (see  \Cref{sec:video_curation}).
Next, we outline the criteria we developed to generate \textbf{ethologically grounded tasks} for canine activity analysis in \Cref{sec:task_design}.
These criteria are developed in close collaboration with an expert canine trainer to ensure coverage of common dog activities and postures.
To scalably generate diverse question-answer pairs, we leverage closed-source VLMs and utilize a \textbf{two-stage QA generation} pipeline (see \Cref{sec:qa_gen}).
We finally implement a \textbf{bias mitigation framework} which leverages an ensemble of frontier LLMs to remove any biases, shortcuts, and hallucinations in the generated QA pairs (see \Cref{sec:bias_mitigation}).

\subsection{Curation of Free-Living Canine Videos from Open Web Sources} \label{sec:video_curation}

We developed a semi-automated pipeline for collecting high-quality, content-specific videos. 
In \texttt{K9-Bench}, videos are curated from publicly available YouTube content, focusing on free-living canines engaged in diverse activities such as interactions with humans, obedience training, and a wide variety of everyday home routines.
Specifically, we implement a video seeding strategy followed by a scalable video retrieval and filtering process (see \Cref{fig:vid_curation_pipeline} for overview).
We provide details below.

\textbf{Initial Video Seeding.}
We first manually curate an initial set of seed videos which will enable scalable programmatic mining of videos. 
We achieve this through: (i) a \textit{query-based search}, where a keyword such as \textit{``Dog Barking"} is used to retrieve relevant videos, and (ii) a \textit{curated seed approach}, where authors compiled videos sourced from open benchmarks \cite{nagrani2024neptune,rawal2024cinepile} or YouTube platform search \footnote{https://developers.google.com/youtube/v3}.
Manual video compilation focused on diverse aspects such as: canine activity occupying majority of the video, non-animated sequences and no harmful content.
We explored animal-centric benchmarks \cite{ng2022animal, liu2023lote, chen2023mammalnet} but found that they do not constitute rich canine-specific video content (see App.
\Cref{tab:animal_datasets_main_paper}).

\textbf{Scalable Video Retrieval Pipeline}
To expand the dataset size beyond the list of manually curated videos, we retrieve additional videos for each seed video via: (a) querying the uploader’s channel, up to $50$ videos, and
(b) the Youtube recommendation engine, through which we collect $5$ videos related to each seed.
This mining strategy ensures systematic video content diversification while ensuring semantic relevance.
Using this video curation pipeline, we mine a set of $1129$ videos.
Statistics on videos collected in this stage are provided in the ``Pre-Refined Filtering Stage" in App. \Cref{tab:video_collection_stats}.

\textbf{Video Filtering Pipeline}
We note that automated video mining retrieves undesired videos for which we design a filtering routine to ensure dataset quality using the closed-source \texttt{Gemini-2.0-flash} VLM.
Text-based filtering is applied to $1129$ videos, which are reduced to $765$ in number (see App. \Cref{sec:initial_text_filtering} for prompt).
This is followed by a $2$-stage manual human review of the $765$ videos to identify issues such as artificially generated content, unnatural dog activities and irrelevant focus which leave us with $498$ videos (see App. \Cref{sec:instruction_guidelines,sec:instruction_guidelines_v2} for verification details).

\textbf{Automated Video Mining using VLM.}
After this, to construct \textit{an automated video rejection pipeline}, we conduct a re-verification of the previous human-verified $765$ videos using the \texttt{Gemini-2.0-flash} model by asking it to reject semantically unrelated videos (see prompt in App. \Cref{appendix:refined_llm_filtering}).
We then compute alignment statistics between the human-based and VLM-based rejection routine, finding a strong correlation between the two.
Specifically, we find that \texttt{Gemini-2.0-Flash} achieves $84.3\%$ binary classification accuracy which quantitatively verifies that it is an effective proxy to conduct automated video rejection on newly mined videos (statistics in App. \Cref{tab:binary_reason_metrics}).
We now repeat the entire video collection and rejection process, this time relying solely on the VLM.
We curate a new seed of $25$ videos from canine YouTube vlog channels which are expanded to $608$ videos using our video retrieval strategy.
We then use \texttt{Gemini-2.0-flash} as the proxy verifier to reject semantically irrelevant videos to obtain a set of $425$ videos.
This gives the final set of $923$ videos, including $498$ videos obtained by manual human rejection.
See ``Post Refined Filtering" in App. \Cref{tab:video_collection_stats} for details.
\vspace{-2mm}

\subsection{Benchmark Task Design}
\label{sec:task_design}

\textbf{Domain-Expert Grounding.}
The \texttt{K9-Bench} task categories are derived from an expert-built canine behavior ontology curated by a certified dog behavior trainer 
(KPA-certified\footnote{https://karenpryoracademy.com/courses/dog-trainer-professional/}).
The ontology contains $70$ canine behavior types annotated with observable visual cues, typical environmental contexts, common daily activities, emotional states, and expert interpretations across breeds and naturalistic settings (see App. \Cref{tab:dog_behaviors} for a subset of documented behaviors).
We then link these identified canine ethological cues to \texttt{K9-Bench} task categories and show representative examples in
App. \Cref{tab:ethogram_compact_rebuttal}.
For example, play bow, tail wagging, prey bow, and pacing are grounded in prior work on canine social signaling, lateralized tail wagging, arousal/aggression cues, and repetitive movement patterns \cite{Byosiere2016,Bekoff1995,Quaranta2007,Siniscalchi2018,Gahwiler2020,Denham2014}.
Where applicable, facial and ear-related cues are further mapped to DogFACS \cite{Waller2013DogFACS}, providing a structured connection between the expert annotations and established canine behavioral analysis.
The resulting cue--context mappings motivate various task categories defined as follows:
\vspace{-3mm}

\begin{enumerate}\setlength{\itemsep}{0.1em}
    \item \textbf{Posture Analysis.}
    Recognizes coarse poses such as sitting, standing, and lying, as well as fine-grained cues including ear orientation, head tilt, body lowering, tail position, gaze direction, and posture changes across frames.

    \item \textbf{Action Sequence.}
    Requires short-horizon temporal reasoning over ordered action transitions, including gait, posture, movement direction, repeated motion, and general behavioral state.

    \item \textbf{Context Analysis.}
    Interprets changes in canine activity and posture in response to environmental stimuli, unfamiliar objects, spatial constraints, human actions, other animals, or object interactions across disjoint temporal windows.

    \item \textbf{Cause--Effect Analysis.}
    Links a query event to subsequent changes in body posture, actions, or interaction patterns across multiple video segments, requiring causal reasoning over temporally separated evidence.

    \item \textbf{Interaction Analysis.}
    Interprets dog--human and dog--dog interactions through visual changes in posture and actions associated with coarse non-verbal communication cues including comfort seeking, play invitations, avoidance or attention seeking.
\end{enumerate}
\vspace{-3mm}

Although \texttt{K9-Bench} is grounded in expert-curated canine behavior knowledge, it does not aim to exhaustively model the full complexity of canine behavior~\cite{bradshaw2016dog,hettsfundamentals}.
Instead, it focuses on common, visually descriptive, and discernible behaviors in naturalistic videos.
Accordingly, the ontology serves as a grounding mechanism for benchmark construction rather than a complete ethological taxonomy.

\textbf{Knowledge Base Creation.}
Because the full ontology contains many fine-grained behavior cues, using it directly as an in-context prompt overloads the VLM during QA generation.
We therefore cluster the ontology, in consultation with the canine expert, into a semantically coherent \emph{Knowledge Base} that supports effective QA generation over long-video.
See App.~\Cref{sec:knowledge_base} for details.

\subsection{Video Question-Answer Pair Generation}
\label{sec:qa_gen}

Recent works show that VLMs can generate synthetic animal-centric question–answer datasets through automated pipelines \cite{jing2024animal,jing2025animal,aljovic2025autonomous,xu2025mousegpt,sun2024video,mamooler2025fine}. Building on this line of work, we construct a canine-centric video question–answer dataset from the $923$ curated videos described in \Cref{sec:video_curation} by leveraging VLMs.
To generate QA pairs grounded in visual analysis, it is essential to capture diverse and meaningful canine states for which we implement a two-stage pipeline used in prior works that curate video understanding datasets \cite{chandrasegaran2024hourvideo,lin2025unleashing,han2025videoespresso}: (i) correct QA generation and (ii) plausible distractor answer generation, as illustrated in Figure~\ref{fig:qa_generation}.
We provide details below.

\textbf{Correct QA Generation.} To ensure that the VLM focuses on generating canine-centric QA pairs, it must be provided with contextual information that guides the generation of relevant and accurate QA pairs.
For this, we leverage the structrued \textit{Knowledge Base} constructed in \Cref{sec:task_design} as the in-context prompt for QA generation in App. \Cref{fig:knowledge_base}.
To generate the correct QA pairs, each video (and audio) is directly supplied to the VLM.
Prompting details are provided in App. \Cref{appendix:qa_gen}.

\textbf{Plausible Wrong Answer Generation}
After generating the correct QA pairs, plausible wrong answers are created to form a $5$-way MCQ dataset.
To achieve this, we first produce detailed textual video narrations which are used as inputs for distractor generation \textit{instead of supplying full videos} in the model context.
This ensures that generated wrong options are not overly specific in their content and have reduced hallucinations.
Each video is processed into temporally segmented narrations that describe scene transitions, spatial layout, canine posture changes and interactions.
These narrations serve as the contextual backbone for generating \textit{grounded} plausible wrong answers (see App. \Cref{fig:video_narration,fig:narration_example} for narration prompt and example output).
For every correct QA pair, we generate $4$ distractor options by prompting the model to produce semantically challenging distractors that remain faithful to canine-centric information in the video while avoiding superficial or repetitive phrasing (see App. \Cref{fig:wrong_answer_gen_prompt} for prompt).
In total, we obtained $8263$ QA pairs across $923$ videos.
\vspace{-1mm}

\subsection{Bias Mitigation Pipeline}  \label{sec:bias_mitigation}

While we leverage \texttt{Gemini-2.5-Flash} for QA generation due to its ability to process long-form videos, we acknowledge that relying solely on a single VLM may introduce strong biases in the generated QA pairs.
Specifically, some generated questions may be solvable without requiring true video understanding: by exploiting prior world knowledge, language-only shortcuts, or speaker-specific cues present in the question phrasing. 
To systematically remove biases arising from \texttt{Gemini} models, we implement an additional bias mitigation pipeline using $4$ distinct frontier LLMs (see \Cref{fig:bias_mitigation_pipeline}) comprising of the following steps.

\textbf{Deaf-Blind LLM Filtering.}  
Following the procedure outlined in \cite{rawal2024cinepile}, we remove questions that are answered correctly without access to video frames.
Specifically, we employ an ensemble of $3$ text-based frontier LLMs \cite{guo2025deepseek,yang2025qwen3,mistral_ai_mistral_small_3_2} from diverse families that are evaluated on the generated QA pairs \textit{without access to video frames}.
Majority voting is done over responses of the $3$ models to filter out QA pairs that are answered correctly, removing $42.58\%$ of the existing QA pool, leading to $4744$ QAs and $907$ videos.
See App. \Cref{appendix:deafblind} for details and App. \Cref{fig:k9bench_appendix_statistics} for dataset stats.

\textbf{Speaker Information Removal.}
Through a qualitative analysis of the filtered QA pairs, we observed that many QA pairs included speaker-specific identifiers (e.g., ``the man,’’ ``the woman,’’ ``speaker 1’’) or auditory cues which lead to easily solving the QAs without requiring actual reasoning over the video frames (see App. \Cref{fig:qa_example}).
To eliminate such shortcuts, we leverage \texttt{GLM-4.5-Flash} \cite{zeng2025glm} to systematically strip speaker identities, auditory references, and overly verbose phrasing from both the questions and answer text.
See App. \Cref{appendix:speaker_information_removal} for prompt details and example rephrasing.
\vspace{-4mm}

\begin{table*}[t]
\centering
\small
\setlength{\tabcolsep}{4.5pt}
\renewcommand{\arraystretch}{1.0}

\begin{tabular}{l c|c| c c c c c}
    \toprule
    Models & Rank & Avg. &
    \cellcolor{cbf!20}\makecell{Posture\\Analysis} &
    \cellcolor{cbf!20}\makecell{Action\\Sequence} &
    \cellcolor{cspu!20}\makecell{Context\\Analysis} &
    \cellcolor{cspu!20}\makecell{Cause-Effect\\Analysis} &
    \cellcolor{csrd!20}\makecell{Interaction\\Analysis}  \\
    \midrule

    \rowcolor{orange!20}
    \multicolumn{8}{c}{\textit{Closed-source Models (API-Based)}} \\

    \texttt{Gemini-2.5 Pro$^\dagger$} & 1 & 40.1 & 37.7 & 36.0 & 44.0 & 44.7 & 42.3   \\
    \texttt{Qwen3VL-235B-A22B$^\dagger$} & 2 & 36.5 & 34.2 & 33.5 & 38.4 & 39.9 & 39.4  \\
    \texttt{GPT-4o$^*$}  & 3 & 30.8 & 29.6 & 26.3 & 35.0 & 33.9 & 33.9  \\

    \rowcolor{pink!20}
    \multicolumn{8}{c}{\textit{Open-source Models$^*$}} \\

    \texttt{Qwen3-VL-4B} & 6 & 27.6 & 27.6 & 24.1 & 30.8 & 28.2 & 30.9  \\
    \texttt{Qwen3-VL-8B} & 5 & 28.8 & 27.8 & 26.5 & 32.1 & 27.6 & 31.9   \\
    \texttt{Qwen3-VL-32B} & 3  & 31.4 & 28.2 & 29.3 & 35.4 & 31.1 & 35.0  \\
    \texttt{InternVL-8B} & 2  & 32.2 & 29.0 & 27.9 & 34.0 & 36.0 & 37.5   \\
    \texttt{InternVL-14B} & 1 & 33.0 & 31.5 & 30.3 & 33.7 & 35.9 & 36.1 \\
    \midrule
    \texttt{Qwen3-VL-32B-Thinking} & 4 & 30.7 & 27.8 & 28.8 & 33.7 & 32.8 & 32.1  \\
    \texttt{Qwen3-VL-8B-Thinking} & 7 & 25.3 & 24.0 & 21.8 & 29.7 & 24.9 & 29.3  \\
    \texttt{Qwen3-VL-4B-Thinking} & 8 & 22.4 & 21.2 & 18.0 & 28.0 & 23.1 & 26.2   \\

    \bottomrule
\end{tabular}

\vspace{-1mm}
\caption{
\textbf{MCQ evaluation results on \texttt{K9-Bench}:} Closed-source models lead overall, while open-source models show varied strengths, particularly in interaction and context reasoning.
$^{*}$ indicates models evaluated with 32 frames. 
$^\dagger$ indicates models evaluated using a 1 FPS setting.
}
\label{tab:canine-results}
\vspace{-4mm}
\end{table*}

\section{Experimental Setup} \label{sec:exp}

\textbf{Model Evaluation Details.} We evaluate a suite of leading closed- and open-source vision–language models on \texttt{K9-Bench}, including \texttt{Gemini 2.5 Pro} \cite{comanici2025gemini}, \texttt{GPT-4o} \cite{hurst2024gpt}, \texttt{Qwen3-VL} \cite{yang2025qwen3}, and  \texttt{InternVL-3.5} \cite{wang2025internvl3} 
(Table~\ref{tab:canine-results}). 
For all open source model and GPT-4o experiments, models receive 32 uniformly sampled video frames and text question, whereas for \texttt{Gemini-2.5-Pro}, \texttt{Qwen3-Omni-Flash} and \texttt{Qwen3VL-235B-A22B}, we provide full-length videos at 1 FPS.
We additionally evaluate models equipped with multimodal input capability (audio and video) such as \texttt{Gemini-2.5 Pro} and \texttt{Qwen3-Omni-Flash} (see Table~\ref{tab:k9bench-multimodal-combined}). 
Since \texttt{Qwen3-Omni-Flash} only supports up to $150$s video and audio processing, the dataset was filtered to $2693$ QA pairs where video length is $<150s$.
Prompts follow the format in App. \Cref{fig:mllm_eval_prompt} and models are instructed to output a free-form answer.
For all thinking model variants, we use a maximum response limit of $2048$ tokens.
Compute details used for open-source model experimentation are provided in App.~\Cref{sec:computational_setup}.

\textbf{MCQ-based Evaluations.} We follow a two-stage criterion for free-form response evaluation similar to \cite{han2025videoespresso}. We first compute the cosine similarity between the embeddings of the probed model’s generated response and each candidate option, using text embeddings from the \texttt{Qwen3-Embeddings-8B} model \cite{zhang2025qwen3}.
Model response is treated as \textit{potentially correct} if its cosine similarity with the ground-truth option exceeds $0.5$.
After this initial check, the response is deemed correct only if no incorrect option attains a higher cosine similarity with the model response than the ground-truth option.

\textbf{Subjective Evaluations.}
We also present LLM-as-a-Judge evaluations (\texttt{GPT-4o} as judge) in addition to the aforementioned MCQ-based evals.
We evaluate the free-form answer from VLMs across five dimensions: logical consistency, factual correctness, accuracy, conciseness, and overall response quality. 
Each dimension is scored on a scale from $0-10$ by \texttt{GPT-4o}. 
We further report aggregated metric scores for each model, normalized to a $0$–$100$ scale, in \Cref{tab:k9bench-llm-as-judge}.
The evaluation criteria and prompts used for the \texttt{GPT-4o} judge are provided in App. \Cref{appendix:subjective_eval}.

\begin{table*}[t]
\centering
\footnotesize
\setlength{\tabcolsep}{2.2pt}
\renewcommand{\arraystretch}{0.92}

\resizebox{\textwidth}{!}{
\begin{tabular}{l | c c c c c c | c c c c c}
\toprule
\multirow{2}{*}{Models}
&
\multicolumn{6}{c|}{Multiple-Choice Accuracy (\%)} 
&
\multicolumn{5}{c}{Subjective Evaluation} \\
\cmidrule(lr){2-7}
\cmidrule(lr){8-12}
&
\cellcolor{cbf!20}\makecell{Posture\\Analysis} &
\cellcolor{cbf!20}\makecell{Action\\Sequence} &
\cellcolor{cspu!20}\makecell{Context\\Analysis} &
\cellcolor{cspu!20}\makecell{Cause-Effect\\Analysis} &
\cellcolor{csrd!20}\makecell{Interaction\\Analysis} &
Avg.
&
\makecell{Logical} &
\makecell{Factual} &
\makecell{Concise} &
\makecell{Accuracy} &
\makecell{Overall} \\
\midrule

\texttt{Gemini-2.5 Pro (A+V)} & 35.6 & 38.3 & 44.8 & 49.6 & 44.6 & 41.9
& 77.41 & 69.98 & 79.57 & 64.97 & 70.81 \\

\texttt{Gemini-2.5 Pro}       & 37.7 & 36.0 & 44.0 & 44.7 & 42.3 & 40.1 & 75.9 & 68.33 & 78.92 & 63.18 & 69.15 \\

\midrule

\texttt{Gemini-2.5 Pro (A+V)}
& 38.44 & 37.82 & 45.15 & 50.0 & 45.04 & 42.37
& 77.94 & 71.13 & 80.41 & 65.77 & 71.54 \\

\texttt{Qwen3-Omni-Flash (A+V)}
& 36.5 & 33.7 & 37.9 & 42.3 & 40.2 & 37.3
& 73.82 & 67.62 & 82.11 & 60.56 & 67.92 \\

\bottomrule
\end{tabular}
}

\vspace{-1mm}
\caption{
\textbf{\texttt{K9-Bench} evaluations using multimodal audio-video inputs}.
Top rowset indicates evaluation on full dataset.
Bottom rowset indicates performance on $2693$ VQAs with video length $<150$s since \texttt{Qwen3-Omni-Flash} supports $<150$s video lengths. All evaluations reported with $1$ FPS.
}
\label{tab:k9bench-multimodal-combined}
\vspace{-6mm}
\end{table*}

\section{Results and Analysis} \label{sec:results}

\vspace{-2mm}
\textbf{Closed Models Performance.}
As shown in \Cref{tab:canine-results}, current MLLMs demonstrate non-trivial yet limited understanding of the canine activities and human-pet interactions.
The best performing closed source model, \texttt{Gemini-2.5 Pro}, achieves an overall MCQ accuracy of $40.1\%$, followed by \texttt{Qwen3VL-235B-A22B} at $36.5\%$ and \texttt{GPT-4o} at $30.8\%$ (but only processes $32$ frames).
Examining performance across task categories reveals consistent trends, with cause-effect analysis being among the strongest categories for top models ($44.7\%$ for \texttt{Gemini-2.5 Pro}) likely because models can use high-level semantic priors to infer plausible causal relationships from coarse visual context.
In contrast, action sequence and posture analysis are the hardest categories across all closed-source models ($36.0\%$ and $37.7\%$ respectively for \texttt{Gemini-2.5 Pro}), as they require precise temporal decomposition of fine-grained behavioral cues, a capability that requires dense frame-level grounding.
Interaction analysis and context analysis have higher performance, indicating that reasoning over social and contextual cues is easier than strict action sequencing and posture analysis.

\textbf{Open Models Performance.}
Among open-source models (\Cref{tab:canine-results}), performance is consistently lower than that of the best closed-source systems.
\texttt{InternVL-14B} leads the open-source group at $33.0\%$, followed closely by \texttt{InternVL-8B} ($32.2\%$) and \texttt{Qwen3-VL-32B} ($31.4\%$). 
Smaller \texttt{Qwen3-VL-8B} and \texttt{Qwen3-VL-4B} models perform worse, achieving $28.8\%$ and $27.6\%$ respectively.
Performance across task categories mirrors trends observed in closed-source models where action sequence and posture analysis are the most challenging; whereas interaction, context and cause-effect analysis maintain higher performance.

\begin{wraptable}{r}{0.56\textwidth}
\vspace{-4mm}
\centering
\scriptsize
\setlength{\tabcolsep}{2pt}
\renewcommand{\arraystretch}{0.9}

\begin{tabular}{l | c c c c c}
\toprule
Models &
\makecell{Logical} &
\makecell{Factual} &
\makecell{Concise} &
\makecell{Accuracy} &
\makecell{Overall} \\

\midrule
\rowcolor{orange!20}
\multicolumn{6}{c}{\textit{Closed-source Models (API-Based)}} \\

\texttt{Gemini-2.5 Pro$^{\dagger}$}  & 75.9 & 68.33 & 78.92 & 63.18 & 69.15 \\
\texttt{Qwen3VL-235B-A22B$^{\dagger}$}  & 74.29 & 67.07 & 81.40 & 60.78 & 67.96 \\
\texttt{GPT-4o$^*$} & 65.96 & 61.90 & 81.34 & 52.81 & 60.99 \\

\midrule
\rowcolor{pink!20}
\multicolumn{6}{c}{\textit{Open-source Models$^*$}} \\

\texttt{Qwen3-VL-8B} & 57.50 & 52.86 & 81.56 & 44.42 & 52.64 \\
\texttt{Qwen3-VL-32B} & 61.94 & 56.20 & 79.90 & 47.60 & 55.73 \\
\texttt{InternVL-8B} & 68.80 & 63.09 & 81.52 & 54.49 & 62.85 \\
\texttt{InternVL-14B} & 68.29 & 63.26 & 82.10 & 54.31 & 62.70 \\
\texttt{Qwen3-VL-32B-Thinking} & 65.67 & 60.01 & 79.97 & 52.52 & 60.18 \\
\texttt{Qwen3-VL-8B-Thinking} & 68.63 & 63.41 & 80.26 & 46.21 & 63.27 \\

\bottomrule
\end{tabular}

\caption{
\textbf{Subjective evaluation} results on \texttt{K9-Bench}. 
$^{*}$ indicates models evaluated with 32 frames. 
$^\dagger$ indicates models evaluated using a 1 FPS setting.
 }
\label{tab:k9bench-llm-as-judge}
\vspace{-4mm}
\end{wraptable}

\textbf{Subjective Evals Analysis.}
\Cref{tab:k9bench-llm-as-judge} reports LLM-as-a-Judge scores across five criteria.
The ranking broadly aligns with MCQ evaluation for closed-source models, with \texttt{Gemini-2.5 Pro} leading at $69.15$, followed by \texttt{Qwen3VL-235B-A22B} and \texttt{GPT-4o}; among open-source models, however, \texttt{Qwen3-VL-8B-Thinking} ranks highest at $63.27$, diverging from its lower MCQ standing, which we discuss further in thinking model analysis below.
Among non-thinking open-source models, \texttt{InternVL} variants outperform all \texttt{Qwen3-VL} counterparts.
Notably, conciseness scores are high and relatively uniform across all models ($\approx$$79$-$82$), indicating that response verbosity is not a primary differentiator; rather the performance gaps are driven by accuracy and factual scores, where closed-source models dominate.
See App. \Cref{fig:subjective_eval_analyses} for sample score assignments.

\begin{figure}[t]
\centering

\begin{subfigure}{0.32\columnwidth}
\centering
\includegraphics[width=\linewidth]{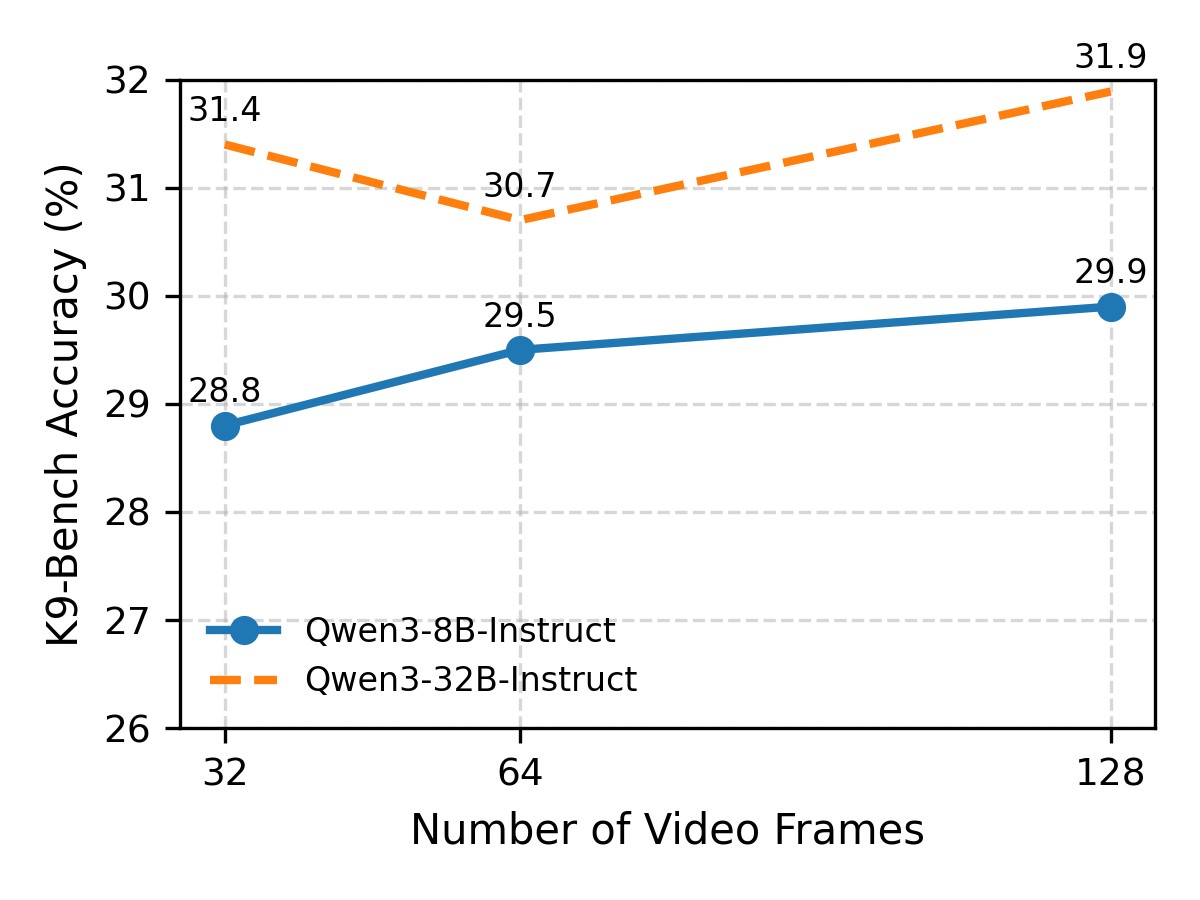}
\vspace{-2mm}
\caption{}
\label{fig:frames_ablation}
\end{subfigure}
\hfill
\begin{subfigure}{0.32\columnwidth}
\centering
\includegraphics[width=\linewidth, trim=400 150 400 150, clip]{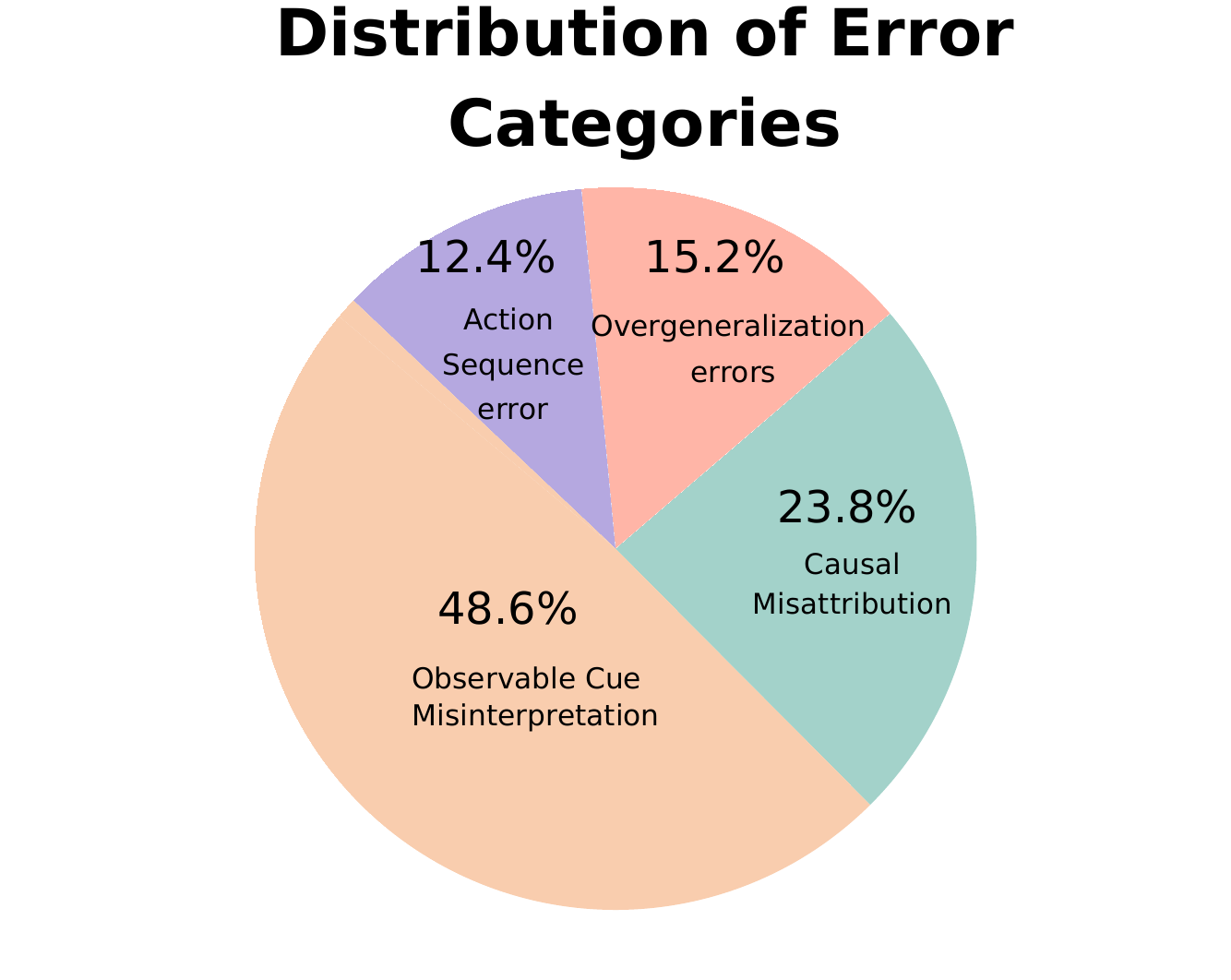}
\caption{}
\label{fig:failure_analysis}
\end{subfigure}
\hfill
\begin{subfigure}{0.32\columnwidth}
\centering
\includegraphics[width=\linewidth]{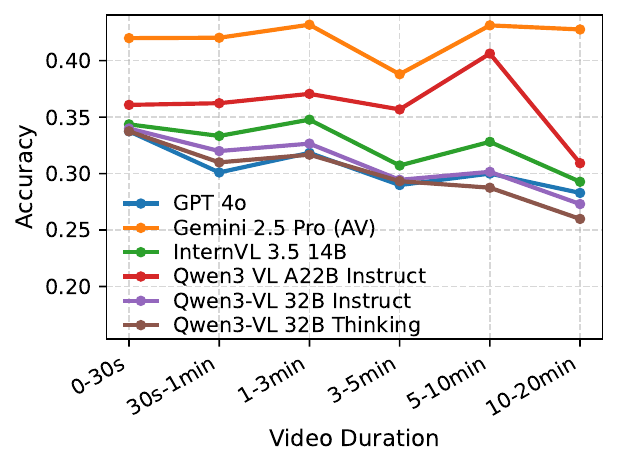}
\caption{}
\label{fig:acc_duration}
\end{subfigure}
\vspace{-2mm}
\caption{\textbf{(a)} Temporal resolution analysis based on MCQ Evals. \textbf{(b)} Breakdown of model response failure types on $105$. \textbf{(c)} Accuracy vs video duration.}
\vspace{-4mm}
\label{fig:k9bench_analysis}
\end{figure}

\textbf{Thinking vs Non-Thinking Models.} 
We find mixed results comparing \texttt{Thinking} variants to their non-\texttt{Thinking} counterparts across both evaluation protocols (\Cref{tab:canine-results,tab:k9bench-llm-as-judge}).
In MCQ evaluation, \texttt{Qwen3-VL-32B-Thinking} ($30.7\%$) achieves lower accuracy than \texttt{Qwen3-VL-32B} ($31.4\%$), and \texttt{Qwen3-VL-8B-Thinking} ($25.3\%$) shows a similar decline relative to the instruct model ($28.8\%$).
In subjective evaluation, however, the trend reverses with \texttt{Qwen3-VL-32B-Thinking} achieving a higher overall score ($60.18$ vs.\ $55.73$) and \texttt{Qwen3-VL-8B-Thinking} similarly improves ($63.27$ vs.\ $52.64$), yet their accuracy scores remain comparable or lower ($52.52$ vs.\ $47.60$ and $46.21$ vs.\ $44.42$ respectively).
This indicates that chain-of-thought reasoning improves response coherence and perceived logical quality but does not consistently improve behavioral grounding.
App. \Cref{fig:qualitative_think_c} illustrates this behavior, where the thinking model over-analyzes visual input and loses focus on the core question, whereas the non-thinking model produces a more direct and accurate response.

\textbf{Do Multimodal Inputs Help?}
We equip \texttt{Gemini-2.5 Pro} with audio inputs to probe the effect of richer multimodal input (\Cref{tab:k9bench-multimodal-combined}).
Adding audio leads to a modest improvement in MCQ accuracy, with overall accuracy increasing from $40.1\%$ to $41.9\%$.
We observe improvements in action sequence analysis ($+2.3$ pp), the hardest category in our benchmark, where audio provides temporal landmarks (such as barks, whines and contact sounds) that make event boundaries more salient which helps the model track ordering of sub-events.
In subjective evaluation, \texttt{Gemini-2.5 Pro (A+V)} scores higher than the video-only version on both accuracy ($64.97$ vs.\ $63.13$) and overall ($70.81$ vs.\ $69.15$) metrics, suggesting that the gains reflect genuine improvements in response quality rather than stylistic differences.
Furthermore, \texttt{Qwen3-Omni-Flash (A+V)} achieves $37.3\%$ indicating that frontier open-source variants lag behind closed models when supplied with multiple modalities in the input context.
App. \Cref{fig:qualitative_human_b} illustrates a case where the audio+video model leverages additional audio cues to generate a more accurate answer with higher accuracy in comparison to the video-only counterpart.

\begin{wraptable}{r}{0.4\columnwidth}
\vspace{-3mm}
\centering
\tiny
\setlength{\tabcolsep}{3pt}
\renewcommand{\arraystretch}{1.0}

\resizebox{\linewidth}{!}{
\begin{tabular}{l c}
\toprule
Models & Accuracy \\
\midrule

\texttt{Qwen3VL-32B}  & 48.4 \\ 
\texttt{InternVL-3.5-14B}  & 54.3 \\ 
\texttt{Gemini-2.5 Pro (A+V)}  & 63.9 \\ 
\texttt{\textbf{Human}}  & \textbf{69.9} \\

\bottomrule
\end{tabular}
}

\caption{\textbf{Human- evaluation} results on $100$ QA subset from \texttt{K9-Bench}.}
\label{tab:human-llm-as-judge}
\vspace{-4mm}
\end{wraptable}

\textbf{Human Evaluation.}
To assess human performance on \texttt{K9-Bench}, $3$ independent human evaluators provided free-form answers to $100$ VQA instances, which are scored using the same LLM-as-a-Judge criteria (App. \Cref{fig:subjective_eval_prompt}) applied to models (\Cref{tab:human-llm-as-judge}).
Human evaluators achieve an accuracy of $69.9\%$, substantially outperforming all evaluated models including \texttt{Gemini-2.5 Pro (A+V)} ($63.9\%$) and the best open-source model \texttt{InternVL-3.5-14B} ($54.3\%$).
Evaluators were instructed to ground their responses in concrete observable evidence while avoiding unsupported inferences about intent or emotion (see GUI for human evaluations in App. \Cref{fig:human_gui}).
This gap is primarily attributable to human ability to integrate relevant visual evidence and audio cues over longer temporal horizon. 
See App. \Cref{fig:qualitative_human_a} for examples.

\textbf{Effect of Temporal Resolution.}
We ablate the number of input frames for \texttt{Qwen3-VL-8B} and \texttt{Qwen3-VL-32B}  (\Cref{fig:frames_ablation}).
Increasing clip length from $32$ to $64$ frames yields small gain for 8B ($+0.7\%$) and slight drop for 32B ($-0.7\%$).
Extending to $128$ frames provides only marginal improvements, reaching $29.9\%$ for 8B and $31.9\%$ for 32B. 
Overall, longer temporal context offers limited benefits with performance largely saturating beyond $64$ frames, suggesting that simply adding more frames is not sufficient and better temporal modeling is needed for larger improvements.
We also report MCQ accuracy as a function of video duration (\Cref{fig:acc_duration}), where video length distribution across VQAs is provided in App. \Cref{fig:vqa_dist}. 
\texttt{Gemini-2.5 Pro (A+V)} maintains consistently higher accuracy across duration bins whereas other models show degradation from $3$-$5$ min range onwards.

\textbf{Failure Mode Analysis.} 
We diagnose the main bottlenecks in the best-performing MLLM, \texttt{Gemini-2.5 Pro (A+V)}, by manually analyzing errors across $105$ VQA instances, categorizing them into four failure types (see App. \Cref{appendix:qualitative_analyses} for definitions).
\Cref{fig:failure_analysis} shows that \textit{Observable Cue Misinterpretation} is the dominant failure mode, accounting for $48.0\%$ of errors.
MLLMs often overlook or misclassify subtle cues such as eye, ear, and tail movements, with irrelevant details further obscuring these signals.
This underscores the difficulty of grounding nuanced canine postural information in videos.
See App. \Cref{appendix:qualitative_analyses} for failure mode examples.

\vspace{-3mm}
\section{Conclusion}
\vspace{-2mm}
In summary, we present \texttt{K9-Bench}, a canine-centric benchmark of 4744 QAs across 907 videos spanning five long-form video tasks built via a scalable VLM/LLM-driven pipeline to evaluate fine-grained, multimodal reasoning in MLLMs. Our experiments show that both open- and closed-source frontier models struggle with compositional reasoning over subtle, temporally extended canine interactions, and that generic chain-of-thought prompting yields only limited gains for such long-horizon scenarios. While our dataset remains comparatively smaller in scale than large vision–language corpora, \texttt{K9-Bench} offers a practical, extensible framework for constructing video reasoning benchmarks in low-data regimes.
Limitations are provided in App.~\Cref{sec:limitations}.

{
    \small
    \bibliographystyle{unsrt}   %
    \bibliography{main}

@String(CVPR= {IEEE Conf. Comput. Vis. Pattern Recog.})

@String(CVPR  = {CVPR})

@inproceedings{gabeff2025mammalps,
  title={MammAlps: A multi-view video behavior monitoring dataset of wild mammals in the Swiss Alps},
  author={Gabeff, Valentin and Qi, Haozhe and Flaherty, Brendan and Sumbul, Gencer and Mathis, Alexander and Tuia, Devis},
  booktitle={Proceedings of the Computer Vision and Pattern Recognition Conference},
  pages={13854--13864},
  year={2025}
}

@inproceedings{ng2022animal,
  title={Animal kingdom: A large and diverse dataset for animal behavior understanding},
  author={Ng, Xun Long and Ong, Kian Eng and Zheng, Qichen and Ni, Yun and Yeo, Si Yong and Liu, Jun},
  booktitle={Proceedings of the IEEE/CVF conference on computer vision and pattern recognition},
  pages={19023--19034},
  year={2022}
}

@inproceedings{chen2023mammalnet,
  title={Mammalnet: A large-scale video benchmark for mammal recognition and behavior understanding},
  author={Chen, Jun and Hu, Ming and Coker, Darren J and Berumen, Michael L and Costelloe, Blair and Beery, Sara and Rohrbach, Anna and Elhoseiny, Mohamed},
  booktitle={Proceedings of the IEEE/CVF conference on computer vision and pattern recognition},
  pages={13052--13061},
  year={2023}
}

@inproceedings{liu2023lote,
  title={LoTE-Animal: A long time-span dataset for endangered animal behavior understanding},
  author={Liu, Dan and Hou, Jin and Huang, Shaoli and Liu, Jing and He, Yuxin and Zheng, Bochuan and Ning, Jifeng and Zhang, Jingdong},
  booktitle={Proceedings of the IEEE/CVF international conference on computer vision},
  pages={20064--20075},
  year={2023}
}

@inproceedings{khosla2011novel,
  title={Novel dataset for fine-grained image categorization: Stanford dogs},
  author={Khosla, Aditya and Jayadevaprakash, Nityananda and Yao, Bangpeng and Li, Fei-Fei},
  booktitle={Proc. CVPR workshop on fine-grained visual categorization (FGVC)},
  volume={2},
  year={2011}
}

@misc{gemini_video_understanding,
  title        = {Video understanding | {Gemini} API},
  author       = {{Google AI Developers}},
  howpublished = {\url{https://ai.google.dev/gemini-api/docs/video-understanding}},
  note         = {Accessed: 2025-08-31}
}

@article{lin2025unleashing,
  title={Unleashing Hour-Scale Video Training for Long Video-Language Understanding},
  author={Lin, Jingyang and Wu, Jialian and Sun, Ximeng and Wang, Ze and Liu, Jiang and Su, Yusheng and Yu, Xiaodong and Chen, Hao and Luo, Jiebo and Liu, Zicheng and others},
  journal={arXiv preprint arXiv:2506.05332},
  year={2025}
}

@misc{xu2025mousegpt,
  title={MouseGPT: A Large-scale Vision-Language Model for Mouse Behavior Analysis},
  author={Teng Xu and Taotao Zhou and Youjia Wang and Peng Yang and Simin Tang and Kuixiang Shao and Zifeng Tang and Yifei Liu and Xinyuan Chen and Hongshuang Wang and others},
  year={2025},
  archivePrefix={bioRxiv},
  eprint={2025-03},
  publisher={Cold Spring Harbor Laboratory}
}

@misc{team2024gemini,
  title={Gemini 1.5: Unlocking Multimodal Understanding Across Millions of Tokens of Context},
  author={Gemini Team and Petko Georgiev and Ving Ian Lei and Ryan Burnell and Libin Bai and Anmol Gulati and Garrett Tanzer and Damien Vincent and Zhufeng Pan and Shibo Wang and others},
  year={2024},
  eprint={2403.05530},
  archivePrefix={arXiv},
  primaryClass={cs.CL}
}

@misc{bai2025qwen2,
  title={Qwen2.5-VL Technical Report},
  author={Shuai Bai and Keqin Chen and Xuejing Liu and Jialin Wang and Wenbin Ge and Sibo Song and Kai Dang and Peng Wang and Shijie Wang and Jun Tang and others},
  year={2025},
  eprint={2502.13923},
  archivePrefix={arXiv},
  primaryClass={cs.CL}
}

@misc{nagrani2024neptune,
  title={Neptune: The Long Orbit to Benchmarking Long Video Understanding},
  author={Arsha Nagrani and Mingda Zhang and Ramin Mehran and Rachel Hornung and Nitesh Bharadwaj Gundavarapu and Nilpa Jha and Austin Myers and Xingyi Zhou and Boqing Gong and Cordelia Schmid and others},
  year={2024},
  eprint={2412.09582},
  archivePrefix={arXiv},
  primaryClass={cs.CV}
}

@misc{rawal2024cinepile,
  title={Cinepile: A Long Video Question Answering Dataset and Benchmark},
  author={Ruchit Rawal and Khalid Saifullah and Miquel Farr{\'e} and Ronen Basri and David Jacobs and Gowthami Somepalli and Tom Goldstein},
  year={2024},
  eprint={2405.08813},
  archivePrefix={arXiv},
  primaryClass={cs.CV}
}

@inproceedings{xiao2021next,
  title={Next-qa: Next phase of question-answering to explaining temporal actions},
  author={Xiao, Junbin and Shang, Xindi and Yao, Angela and Chua, Tat-Seng},
  booktitle={Proceedings of the IEEE/CVF conference on computer vision and pattern recognition},
  pages={9777--9786},
  year={2021}
}

@article{mangalam2023egoschema,
  title={Egoschema: A diagnostic benchmark for very long-form video language understanding},
  author={Mangalam, Karttikeya and Akshulakov, Raiymbek and Malik, Jitendra},
  journal={Advances in Neural Information Processing Systems},
  volume={36},
  pages={46212--46244},
  year={2023}
}

@misc{fu2024video,
  title={Video-MME: The First-Ever Comprehensive Evaluation Benchmark of Multi-modal LLMs in Video Analysis},
  author={Chaoyou Fu and Yuhan Dai and Yondong Luo and Lei Li and Shuhuai Ren and Renrui Zhang and Zihan Wang and Chenyu Zhou and Yunhang Shen and Mengdan Zhang and others},
  year={2024},
  eprint={2405.21075},
  archivePrefix={arXiv},
  primaryClass={cs.CV}
}

@misc{zhang2024llavanextvideo,
  title={LLaVA-NeXT: A Strong Zero-shot Video Understanding Model},
  url={https://llava-vl.github.io/blog/2024-04-30-llava-next-video/},
  author={Zhang, Yuanhan and Li, Bo and Liu, haotian and Lee, Yong jae and Gui, Liangke and Fu, Di and Feng, Jiashi and Liu, Ziwei and Li, Chunyuan},
  month={April},
  year={2024}
}

@inproceedings{han2025videoespresso,
  title={Videoespresso: A large-scale chain-of-thought dataset for fine-grained video reasoning via core frame selection},
  author={Han, Songhao and Huang, Wei and Shi, Hairong and Zhuo, Le and Su, Xiu and Zhang, Shifeng and Zhou, Xu and Qi, Xiaojuan and Liao, Yue and Liu, Si},
  booktitle={Proceedings of the Computer Vision and Pattern Recognition Conference},
  pages={26181--26191},
  year={2025}
}

@article{comanici2025gemini,
  title={Gemini 2.5: Pushing the frontier with advanced reasoning, multimodality, long context, and next generation agentic capabilities},
  author={Comanici, Gheorghe and Bieber, Eric and Schaekermann, Mike and Pasupat, Ice and Sachdeva, Noveen and Dhillon, Inderjit and Blistein, Marcel and Ram, Ori and Zhang, Dan and Rosen, Evan and others},
  journal={arXiv preprint arXiv:2507.06261},
  year={2025}
}

@article{hurst2024gpt,
  title={Gpt-4o system card},
  author={Hurst, Aaron and Lerer, Adam and Goucher, Adam P and Perelman, Adam and Ramesh, Aditya and Clark, Aidan and Ostrow, AJ and Welihinda, Akila and Hayes, Alan and Radford, Alec and others},
  journal={arXiv preprint arXiv:2410.21276},
  year={2024}
}

@article{yang2025qwen3,
  title={Qwen3 technical report},
  author={Yang, An and Li, Anfeng and Yang, Baosong and Zhang, Beichen and Hui, Binyuan and Zheng, Bo and Yu, Bowen and Gao, Chang and Huang, Chengen and Lv, Chenxu and others},
  journal={arXiv preprint arXiv:2505.09388},
  year={2025}
}

@article{wang2025internvl3,
  title={Internvl3. 5: Advancing open-source multimodal models in versatility, reasoning, and efficiency},
  author={Wang, Weiyun and Gao, Zhangwei and Gu, Lixin and Pu, Hengjun and Cui, Long and Wei, Xingguang and Liu, Zhaoyang and Jing, Linglin and Ye, Shenglong and Shao, Jie and others},
  journal={arXiv preprint arXiv:2508.18265},
  year={2025}
}

@article{zhang2025qwen3,
  title={Qwen3 Embedding: Advancing Text Embedding and Reranking Through Foundation Models},
  author={Zhang, Yanzhao and Li, Mingxin and Long, Dingkun and Zhang, Xin and Lin, Huan and Yang, Baosong and Xie, Pengjun and Yang, An and Liu, Dayiheng and Lin, Junyang and others},
  journal={arXiv preprint arXiv:2506.05176},
  year={2025}
}

@article{guo2025deepseek,
  title={Deepseek-r1: Incentivizing reasoning capability in llms via reinforcement learning},
  author={Guo, Daya and Yang, Dejian and Zhang, Haowei and Song, Junxiao and Zhang, Ruoyu and Xu, Runxin and Zhu, Qihao and Ma, Shirong and Wang, Peiyi and Bi, Xiao and others},
  journal={arXiv preprint arXiv:2501.12948},
  year={2025}
}

@article{jing2024animal,
  title={Animal-bench: Benchmarking multimodal video models for animal-centric video understanding},
  author={Jing, Yinuo and Zhang, Ruxu and Liang, Kongming and Li, Yongxiang and He, Zhongjiang and Ma, Zhanyu and Guo, Jun},
  journal={Advances in Neural Information Processing Systems},
  volume={37},
  pages={78766--78796},
  year={2024}
}

@article{aljovic2025autonomous,
  title={An autonomous AI agent for universal behavior analysis},
  author={Aljovi{\'c}, Almir and Lin, Zuwan and Wang, Wenbo and Zhang, Xinhe and Marin-Llobet, Arnau and Liang, Ningyue and Canales, Bradley and Lee, Jaeyong and Baek, Jongmin and Liu, Ren and others},
  journal={bioRxiv},
  pages={2025--05},
  year={2025},
  publisher={Cold Spring Harbor Laboratory}
}

@inproceedings{mamooler2025fine,
  title={Fine-tuning Vision-Language Models for Animal Behavior Analysis},
  author={Mamooler, Sepideh and Qi, Haozhe and Gabeff, Valentin and Montariol, Syrielle and Bosselut, Antoine and Mathis, Alexander},
  booktitle={LLM for Scientific Discovery: Reasoning, Assistance, and Collaboration},
  year={2025}
}

@article{jing2025animal,
  title={Animal-CLIP: A Dual-Prompt Enhanced Vision-Language Model for Animal Action Recognition},
  author={Jing, Yinuo and Liang, Kongming and Zhang, Ruxu and Sun, Hao and Li, Yongxiang and He, Zhongjiang and Ma, Zhanyu},
  journal={International Journal of Computer Vision},
  pages={1--21},
  year={2025},
  publisher={Springer}
}

@misc{american2025american,
  title={The American Pet Products Association (APPA) Releases 2025 State of the Industry Report},
  author={American Pet Products Association and others},
  year={2025},
  publisher={Von https://americanpetproducts. org/news/the-american-pet~…}
}

@article{chandrasegaran2024hourvideo,
  title={Hourvideo: 1-hour video-language understanding},
  author={Chandrasegaran, Keshigeyan and Gupta, Agrim and Hadzic, Lea M and Kota, Taran and He, Jimming and Eyzaguirre, Crist{\'o}bal and Durante, Zane and Li, Manling and Wu, Jiajun and Fei-Fei, Li},
  journal={Advances in Neural Information Processing Systems},
  volume={37},
  pages={53168--53197},
  year={2024}
}

@article{sun2024video,
  title={Video foundation models for animal behavior analysis},
  author={Sun, Jennifer J and Zhou, Hao and Zhao, Long and Yuan, Liangzhe and Seybold, Bryan and Hendon, David and Schroff, Florian and Ross, David A and Adam, Hartwig and Hu, Bo and others},
  journal={bioRxiv},
  pages={2024--07},
  year={2024},
  publisher={Cold Spring Harbor Laboratory}
}

@article{zeng2025glm,
  title={Glm-4.5: Agentic, reasoning, and coding (arc) foundation models},
  author={Zeng, Aohan and Lv, Xin and Zheng, Qinkai and Hou, Zhenyu and Chen, Bin and Xie, Chengxing and Wang, Cunxiang and Yin, Da and Zeng, Hao and Zhang, Jiajie and others},
  journal={arXiv preprint arXiv:2508.06471},
  year={2025}
}

@inproceedings{yang2025thinking,
  title={Thinking in space: How multimodal large language models see, remember, and recall spaces},
  author={Yang, Jihan and Yang, Shusheng and Gupta, Anjali W and Han, Rilyn and Fei-Fei, Li and Xie, Saining},
  booktitle={Proceedings of the Computer Vision and Pattern Recognition Conference},
  pages={10632--10643},
  year={2025}
}

@article{bradshaw2016dog,
  title={Dog social behavior and communication},
  author={Bradshaw, John and Rooney, Nicola and Serpell, J},
  journal={The domestic dog: Its evolution, behavior and interactions with people},
  volume={2},
  pages={133--159},
  year={2016},
  publisher={Cambridge University Press Cambridge, MA, USA}
}

@misc{hettsfundamentals,
  author       = {Suzanne Hetts and Dan Estep},
  title        = {Lecture Notes for Fundamentals of Canine Ethology Telecourse},
  institution  = {Animal Behavior Associates},
  year         = {2004},
  url          = {https://www.animalbehaviorassociates.com/ezine/HANDOUTCOMPLETE.pdf},
  note         = {Suzanne Hetts: Ph.D., CAAB, CPDT; Dan Estep: Ph.D., CAAB}
}

@article{wu2024longvideobench,
  title={Longvideobench: A benchmark for long-context interleaved video-language understanding},
  author={Wu, Haoning and Li, Dongxu and Chen, Bei and Li, Junnan},
  journal={Advances in Neural Information Processing Systems},
  volume={37},
  pages={28828--28857},
  year={2024}
}

@inproceedings{li2024mvbench,
  title={Mvbench: A comprehensive multi-modal video understanding benchmark},
  author={Li, Kunchang and Wang, Yali and He, Yinan and Li, Yizhuo and Wang, Yi and Liu, Yi and Wang, Zun and Xu, Jilan and Chen, Guo and Luo, Ping and others},
  booktitle={Proceedings of the IEEE/CVF Conference on Computer Vision and Pattern Recognition},
  pages={22195--22206},
  year={2024}
}

@inproceedings{zhou2025mlvu,
  title={Mlvu: Benchmarking multi-task long video understanding},
  author={Zhou, Junjie and Shu, Yan and Zhao, Bo and Wu, Boya and Liang, Zhengyang and Xiao, Shitao and Qin, Minghao and Yang, Xi and Xiong, Yongping and Zhang, Bo and others},
  booktitle={Proceedings of the Computer Vision and Pattern Recognition Conference},
  pages={13691--13701},
  year={2025}
}

@inproceedings{majumdar2024openeqa,
  title={Openeqa: Embodied question answering in the era of foundation models},
  author={Majumdar, Arjun and Ajay, Anurag and Zhang, Xiaohan and Putta, Pranav and Yenamandra, Sriram and Henaff, Mikael and Silwal, Sneha and Mcvay, Paul and Maksymets, Oleksandr and Arnaud, Sergio and others},
  booktitle={Proceedings of the IEEE/CVF conference on computer vision and pattern recognition},
  pages={16488--16498},
  year={2024}
}

@article{liu2020computer,
  title={A computer vision-based method for spatial-temporal action recognition of tail-biting behaviour in group-housed pigs},
  author={Liu, Dong and Oczak, Maciej and Maschat, Kristina and Baumgartner, Johannes and Pletzer, Bernadette and He, Dongjian and Norton, Tomas},
  journal={Biosystems Engineering},
  volume={195},
  pages={27--41},
  year={2020},
  publisher={Elsevier}
}

@inproceedings{mathis2021pretraining,
  title={Pretraining boosts out-of-domain robustness for pose estimation},
  author={Mathis, Alexander and Biasi, Thomas and Schneider, Steffen and Yuksekgonul, Mert and Rogers, Byron and Bethge, Matthias and Mathis, Mackenzie W},
  booktitle={Proceedings of the IEEE/CVF winter conference on applications of computer vision},
  pages={1859--1868},
  year={2021}
}

@article{duporge2025baboonland,
  title={BaboonLand Dataset: Tracking Primates in the Wild and Automating Behaviour Recognition from Drone Videos: I. Duporge et al.},
  author={Duporge, Isla and Kholiavchenko, Maksim and Harel, Roi and Wolf, Scott and Rubenstein, Daniel I and Crofoot, Margaret C and Berger-Wolf, Tanya and Lee, Stephen J and Barreau, Julie and Kline, Jenna and others},
  journal={International Journal of Computer Vision},
  pages={1--12},
  year={2025},
  publisher={Springer}
}

@misc{mistral_ai_mistral_small_3_2,
  title        = {Mistral Small 3.2 – Mistral AI},
  howpublished = {\url{https://docs.mistral.ai/models/mistral-small-3-2-25-06}},
  month        = jun,
  year         = 2025,
  note         = {Version 25.06, “An update to our previous small model”, released June 2025.},
  author       = {Mistral AI}
}

@article{Byosiere2016,
  author  = {Byosiere, Sarah-Elizabeth and others},
  title   = {Investigating the function of play bows in adult pet dogs ({Canis lupus familiaris})},
  journal = {Behavioural Processes},
  year    = {2016}
}

@article{Bekoff1995,
  author  = {Bekoff, Marc},
  title   = {Play Signals as Punctuation: The Structure of Social Play in Canids},
  journal = {Behaviour},
  year    = {1995}
}

@article{Quaranta2007,
  author  = {Quaranta, Angelo and others},
  title   = {Asymmetric tail-wagging responses by dogs to different emotive stimuli},
  journal = {Current Biology},
  year    = {2007}
}

@article{Siniscalchi2018,
  author  = {Siniscalchi, Marcello and others},
  title   = {Communication in Dogs},
  journal = {Animals},
  year    = {2018}
}

@article{Gahwiler2020,
  author  = {G{\"a}hwiler, Sarah and others},
  title   = {Fear expressions of dogs during New Year fireworks: a video analysis},
  journal = {Scientific Reports},
  year    = {2020},
}

@article{Denham2014,
  author  = {Denham, Hamish D. C. and others},
  title   = {Repetitive behaviour in kennelled domestic dog: stereotypical or not?},
  journal = {Physiology \& Behavior},
  year    = {2014},
}

@manual{Waller2013DogFACS,
  author       = {Waller, Bridget and others},
  title        = {DogFACS: The Dog Facial Action Coding System},
  year         = {2013},
}

@article{xu2025qwen3,
  title={Qwen3-omni technical report},
  author={Xu, Jin and Guo, Zhifang and Hu, Hangrui and Chu, Yunfei and Wang, Xiong and He, Jinzheng and Wang, Yuxuan and Shi, Xian and He, Ting and Zhu, Xinfa and others},
  journal={arXiv preprint arXiv:2509.17765},
  year={2025}
}
}

\newpage
\begin{center}
    \vspace{2em} %
    {\LARGE \textbf{Supplementary Material}}
    \vspace{2em} %
\end{center}%

\section*{Appendix Index}
\addcontentsline{toc}{section}{Appendix Index}

\startcontents[appendix] %
\printcontents[appendix]{l}{1}{}

\appendix

\section{Additional Dataset Statistics}
\label{sec:appendix_dataset_statistics}

\begin{figure}[h]
\centering
\captionsetup{font=scriptsize, labelfont=scriptsize}

\begin{subfigure}[t]{0.48\linewidth}
    \centering
    \includegraphics[width=\linewidth]{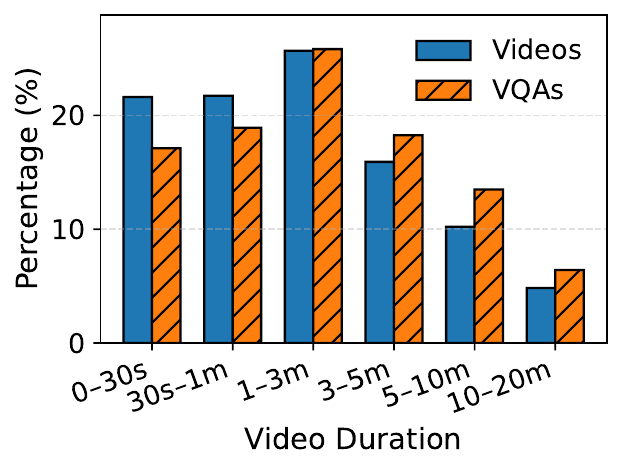}
    \subcaption{VQA distribution across videos}
    \label{fig:vqa_dist}
\end{subfigure}
\hfill
\begin{subfigure}[t]{0.48\linewidth}
    \centering
    \includegraphics[width=\linewidth]{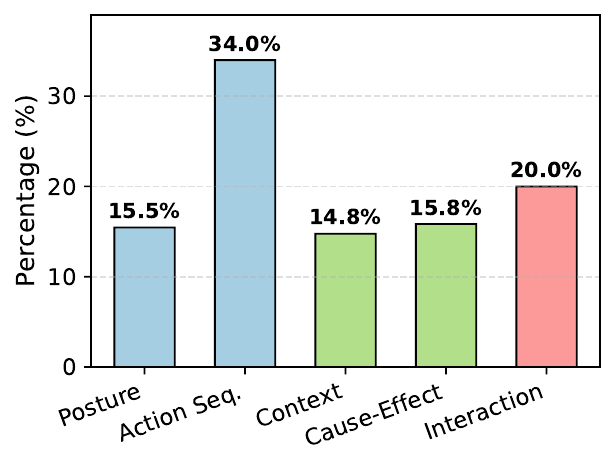}
    \subcaption{Category-wise QA distribution}
    \label{fig:cat_dist}
\end{subfigure}

\vspace{-2mm}

\caption{Distribution of VQAs across videos and category-wise behavior distribution in \texttt{K9-Bench}, illustrating dataset diversity and temporal reasoning complexity.}

\label{fig:k9bench_appendix_statistics}
\vspace{-4mm}
\end{figure}

In this appendix, we provide additional statistical insights into the dataset, including video length distribution, VQA density, and category-level coverage. 
\Cref{fig:k9bench_appendix_statistics} presents the distribution of VQAs across videos and the category-wise behavior distribution, offering a comprehensive overview of dataset diversity and annotation density. 
These statistics provide a deeper understanding of the temporal structure, behavioral coverage, and reasoning complexity of the \texttt{K9-Bench} dataset.

\begin{table*}[t]
\centering
\scriptsize
\begin{tabular}{p{2.2cm} p{1.1cm} p{1.3cm} p{1.4cm} p{4.2cm}}
\toprule
\textbf{Dataset} &
\textbf{Video Samples} &
\textbf{QA Samples} &
\textbf{Canine-Focused QAs} &
\textbf{Tasks} \\
\midrule

LoTE-Animal \cite{liu2023lote}  & $10000$ & $\times$ &  $\times$ & Detection, Segmentation, Action Recognition \\
Animal Kingdom \cite{ng2022animal} & 4301 & $\times$  & $\times$ & Pose Estimation, Action Recognition, Video Grounding \\
MammalNet \cite{chen2023mammalnet}  & $18000$ & $\times$ & $\times$ & Action Recognition \\
Animal-Bench \cite{jing2024animal}  & $31260$ & $\checkmark$ $41839$ & $\checkmark$ 229 QA & Question-Answer, Action Recognition \\
\textbf{K9-Bench (Ours)}  & $907$ & $\checkmark$ $4744$ & $\checkmark$ 4744 QA & Question-Answer, Action Recognition, Context Understanding \\
\bottomrule
\end{tabular}
\caption{Comparison of existing animal datasets with canine-focused question-answering (QA) benchmarks. K9-Bench includes a dedicated set of canine-specific QA samples, allowing systematic evaluation of models on canine-centric video understanding.}
\vspace{-4mm}
\label{tab:animal_datasets_main_paper}
\end{table*}

\vspace{2mm}

\section{Details of Video Collection Pipeline} \label{sec:videocollection}

In this section, we provide specific details of prompting and video filtering pipeline presented in \Cref{sec:video_curation} of main text.

\subsection{Initial LLM Filtering}
\label{sec:initial_text_filtering}

The \textit{Neptune} and \textit{Query - "Dog Barking"} datasets were first filtered using \texttt{Gemini-2.0-Flash} text based filtering before undergoing human validation and expansion of dataset. 

The prompt used for automated filtering is provided in \Cref{fig:video_intial_filtering_prompt}.
This step reduced the dataset from a total of 1,129 videos ($593 + 536$) to 765 videos.

\begin{figure}[ht!]
\centering
\begin{tcolorbox}[title=\texttt{Initial Filtering Prompt}]
\begin{lstlisting}[style=promptstyle,basicstyle=\ttfamily\scriptsize]
Determine if this YouTube video is related to dogs, contains dog information in the video, and is not a compilation of multiple videos, and contains no sexual content:
Title: {video_details['title']}
Description: {video_details['description']}

Respond with 'YES' if it meets the criteria, otherwise 'NO'.
\end{lstlisting}
\end{tcolorbox}
\caption{Initial Video Filtering Prompt.}
\label{fig:video_intial_filtering_prompt}
\end{figure}

\begin{table*}[t]
\centering
\small
\setlength{\tabcolsep}{3.2pt}
\renewcommand{\arraystretch}{1.1}

\begin{tabular}{
p{3cm}
>{\centering\arraybackslash}p{2cm}
>{\centering\arraybackslash}p{2cm}
>{\centering\arraybackslash}p{2cm}
>{\centering\arraybackslash}p{1.6cm}
>{\centering\arraybackslash}p{1.4cm}
}

\toprule
\textbf{Source} & \textbf{Initial Videos} & \textbf{Max Videos per Channel} & \textbf{Max Recommended per Video} & \textbf{Videos Collected (Pipeline)} & \textbf{Filtered Videos} \\ 
\midrule

\multicolumn{6}{l}{\textbf{Pre-Refined Filtering Stage}} \\

Neptune \cite{nagrani2024neptune} & 13 & 50 & 5 & 593 & 261 \\
Query -- ``Dog Barking'' & 10 & 50 & 5 & 536 & 237 \\

\midrule

\multicolumn{6}{l}{\textbf{Post-Refined Filtering Stage}} \\

Dog Vlog Videos & 25 & 100 & 5 & 608 & 425 \\

\bottomrule
\end{tabular}
\caption{Overview of video sources, collection limits, and total videos gathered through the scalable pipeline.}
\label{tab:video_collection_stats}

\end{table*}

\subsection{Human Validation for Filtering Videos} \label{app:human_val_filtering_videos}

The human validation process was performed to filter out the videos and understand valid reasons due to which a video should be rejected. 
This approach enabled the development of clear, consistent filtering guidelines usable by both human reviewers and VLMs in subsequent scalable video filtering. 

\paragraph{Two-Step Human Validation of 765 Videos ($\approx$ 12 Hours Total Footage)}  
The dataset was reviewed in two sequential stages to identify rejection reasons and refine the inclusion and exclusion criteria:  
\begin{itemize}
    \item \textbf{Exploratory Assessment —} We (authors) conducted a rapid pass over the videos ($\approx200$ videos from $765$ videos), noting broad rejection reasons as they occurred. 
    The aim was to list out a coarse category of reasons for rejecting irrelevant videos.
    Frequent issues included: no dog present, artificially generated content, or in appropriate material such as product reviews with minimal dog footage, excessive human discussion, largely inactive dogs with little behavioral context, unsuitable human–dog interactions, duplicate or clipped videos, and static or context-poor footage.
    \item \textbf{Guideline Creation \& Structured Review —} Insights from the previous stage informed the development of inclusion and rejection criteria, as well as a standardized list of video rejection reasons. 
    These guidelines (see \Cref{fig:video_criteria} and \Cref{sec:instruction_guidelines}) were then systematically applied by human reviewer (a professional with over four years of canine-product development experience) on $765$ videos, ensuring consistent and objective filtering.
\end{itemize}

\begin{figure}[!t]
\centering
\begin{minipage}[t]{\textwidth}
\begin{tcolorbox}[
    colback=green!1!white,
    colframe=green!10!gray,
    title={\texttt{Video Inclusion Criteria}},
    boxrule=0.1mm,
    sharp corners,
    enhanced,
    fontupper=\small
    ]
\label{box:video_criteria}
A video is eligible for inclusion if:
\begin{itemize}
    \item The video must contain a real dog.
    \item The dog should be engaged in meaningful activity for a sufficient part of the video and not remain stationary.
\end{itemize}
\end{tcolorbox}

\begin{tcolorbox}[colback=blue!1!white, colframe=blue!10!gray, title=\texttt{Valid Video Rejection Reasons and Criteria}, boxrule=0.01mm, rounded corners, enhanced, fontupper=\small]
For each rejected video, the reviewer must select one or more of the following reasons: \label{box:rejection_reasons_and_criteria}
\begin{itemize}
    \item \textbf{No Dog Present} – The video does not contain any real dog.
    \item \textbf{Artificially Generated} – The dog is generated using generative AI tools (e.g., GPT-based video generation, CGI).
    \item \textbf{Not Appropriate} – The content is inappropriate or unsafe (e.g., abusive behavior toward the dog, middle finger gestures).
    \item \textbf{Stationary or Minimally Active Dog} – The dog is in photograph or banner, inactive for most of the video or only present briefly.
    \item \textbf{Compilation Video} – The video is made up of multiple unrelated clips.
    \item \textbf{Poor Video Quality} – Low resolution, poor lighting, or excessive motion blur prevents meaningful analysis.
    \item \textbf{Irrelevant Focus} – The video focuses on products, scenery, or other subjects rather than the dog's behavior.
    \item \textbf{Unnatural or Staged Scenario} – The behavior or activity is unrealistic or staged in a non-natural environment.
    \item \textbf{Duplicate or Near-Duplicate} – The video or a visually similar video may exist in the dataset, but with a different video ID.
\end{itemize}
\end{tcolorbox}
\end{minipage}

\caption{Video inclusion and rejection reasons and criteria used for dataset filtering and review.}
\label{fig:video_criteria}
\end{figure}

\subsubsection{Instruction Guidelines}
\label{sec:instruction_guidelines}

The authors first reviewed $\approx200$ videos from the $765$ that remain after the initial LLM-based text filtering. 
A detailed guideline for evaluation of the remaining videos is then prepared which summarized reasons that are grounds for video rejection.
These video inclusion and rejection guidelines are presented in \Cref{fig:video_criteria}.
The video inclusion and rejection guidelines are then used to conduct manual review of the $765$ videos.
This is done by an independent professional with over four years of canine-product development experience.

\subsubsection{Human Validation Results}
\label{sec:instruction_guidelines_v2}

\begin{figure}[h]  %
    \centering
    \includegraphics[width=0.7\columnwidth, trim=50mm 3mm 25mm 2mm, clip]{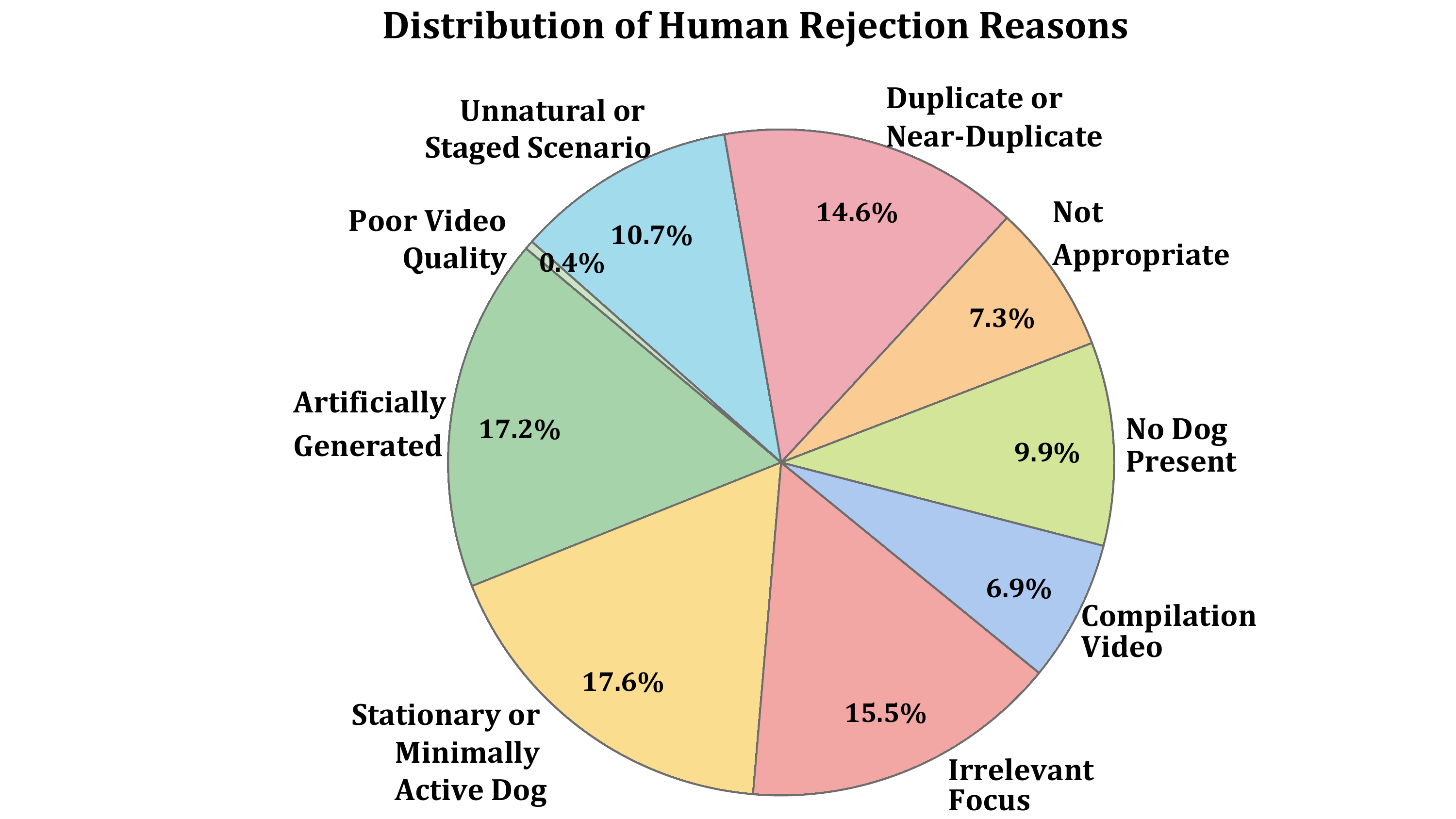}
    \caption{Distribution of rejection reasons for 765 videos reviewed by human annotator. } 
    \label{fig:rejection_summary}
\end{figure}

Following the established guidelines, human reviewer assessed videos and marked them for rejection ($\approx$34.9\%). 
The distribution of rejection reasons for this $765$ videos performed by the human reviewer is visualized in \Cref{fig:rejection_summary}.

\subsection{Refined LLM Filtering} \label{appendix:refined_llm_filtering}

To improve LLM-based filtering, the prompt was aligned with human validation guidelines to capture rejection reasons accurately. Accordingly, up to 512 frames per video were extracted and compiled into a single clip at 1 FPS for \texttt{Gemini-2.0-Flash} processing. 
The \textit{"Duplicate or Near-Duplicate"} category is excluded from the Instruction guidelines in \Cref{fig:video_criteria}, as each video is sent as a separate API request. 
The refined filtering prompt with video modality, provided to \texttt{Gemini-2.0-Flash} is shown in \Cref{fig:refined_filtering_prompt}.

\begin{figure*}[ht!]
\centering
\begin{tcolorbox}[title=\texttt{Refined Filtering Prompt}]
\begin{lstlisting}[style=promptstyle,basicstyle=\ttfamily\scriptsize]
You are reviewing a YouTube video to determine if it should be ACCEPTED or REJECTED for inclusion in a dog behavior dataset.

### Decision Process:
1. Watch the video carefully and associated title and description.
2. Apply the **Inclusion Criteria** and **Rejection Criteria** exactly as listed below.
3. If the video meets ALL Inclusion Criteria and NONE of the Rejection Criteria then Respond with "YES".
4. If the video fails ANY Inclusion Criteria or meets ANY Rejection Criteria then Respond with "NO" and specify EXACTLY ONE OR MORE reasons from the **Allowed Rejection Reasons List**.
---
### Inclusion Criteria (ALL must be true for acceptance):
- The video contains a real dog (not an image, animation, or AI-generated dog).
- The dog is actively engaged in meaningful activity for the majority of the video (not stationary, not appearing only in a photograph, banner, or static presentation).
- The content is natural and realistic, representing genuine pet behavior (vlogs or occasional made-up videos are acceptable if they do not appear overly staged or unrealistic).
- The video is a continuous recording, not a compilation of multiple unrelated clips.
- The video contains no sexual and harmful content, abusive behavior, or inappropriate human actions (e.g., middle finger gestures).
- The video is of sufficient visual quality for analysis (clear resolution, reasonable lighting, no excessive motion blur).
- The dog is a primary focus of the video (not just appearing briefly as a background element or product prop).
---
### Rejection Criteria (ANY of these means rejection):
- No real dog present.
- Dog is artificially generated (e.g. using GPT, CGI, generative AI, animation, presentation slides).
- Video is a compilation of unrelated clips.
- Contains sexual content, abuse toward the dog, or inappropriate gestures by humans.
- Dog is stationary or minimally active for most of the video.
- Very poor visual quality (low resolution, extreme lighting issues, excessive motion blur).
- Dog is not the main focus and does not deliver any meaningful behavior for sufficent amount of time; video focuses on unrelated objects, products, or scenery.
- Depicts unrealistic or staged scenarios.
---
### Allowed Rejection Reasons (choose from this list only):
- No Dog Present
- Artificially Generated
- Compilation Video
- Not Appropriate
- Inappropriate Actions by Humans
- Stationary or Minimally Active Dog
- Poor Video Quality
- Irrelevant Focus
- Unnatural or Staged Scenario
---
### Output Format (JSON):
Return ONLY valid JSON in this exact structure:
{{
  "decision": "YES" or "NO",
  "reasons": []  // If decision is NO, list one or more reasons from the allowed rejection reasons list
}}
---
### Video Metadata:
Title: {title}
Description: {description}
\end{lstlisting}
\end{tcolorbox}
\caption{Refined filtering prompt used for automated video filtering
.}
\label{fig:refined_filtering_prompt}
\end{figure*}

\begin{table*}[!t]
\centering
\scriptsize
\setlength{\tabcolsep}{5pt}
\renewcommand{\arraystretch}{1.0}
\begin{tabular}{p{3cm} *{7}{>{\centering\arraybackslash}p{1.1cm}}}
\toprule
\multirow{2}{*}{\textbf{Rejection Categories}} & \multicolumn{5}{c}{\textbf{Binary Rejection}} & \multicolumn{2}{c}{\textbf{Reason Matching}} \\
\cmidrule(lr){2-6} \cmidrule(lr){7-8}
 & TP & FN & FP & Recall & Precision & Match & Align (\%) \\
\midrule
Stationary / Minimally Active & 24 & 23 & 5 & 51.1 & 82.8 & 13 & 54.2 \\
Artificially Generated & 38 & 8  & 0 & 82.6 & 100.0 & 17 & 44.7 \\
Irrelevant Focus & 23 & 18 & 8 & 56.1 & 74.2 & 22 & 95.7 \\
Duplicate / Near-Duplicate & 11 & 28 & 0 & 28.2 & 100.0 & 0 & 0.0 \\
Unnatural / Staged Scenario & 24 & 5 & 8 & 82.8 & 75.0 & 24 & 100.0 \\
No Dog Present & 23 & 3 & 0 & 88.5 & 100.0 & 8 & 34.8 \\
Not Appropriate & 14 & 5 & 1 & 73.7 & 93.3 & 6 & 42.9 \\
Compilation Video & 14 & 4 & 3 & 77.8 & 82.4 & 8 & 57.1 \\
Poor Video Quality & 1 & 0 & 0 & 100.0 & 100.0 & 0 & 0.0 \\
\midrule
\textbf{Total} & 172 & 95 & 25 & 64.4 & 87.3 & 106 & 61.6 \\
\bottomrule
\end{tabular}%
\caption{LLM vs human rejection of initial video rejection on a 765-video subset used in \Cref{fig:rejection_summary} followed by an additional LLM run using a refined filtering prompt (see \Cref{fig:refined_filtering_prompt}). We report binary rejection metrics (True Positives, False Negatives, False Positives, Recall, and Precision) and reason-level alignment between LLM and human, highlighting varying agreement across rejection categories such as synthetic, duplicate, and non-canine videos.} \label{tab:binary_reason_metrics}
\end{table*}

\paragraph{Gemini-2.0-Flash - Video Rejection Performance Analyses:}
Following human validation, we evaluated the refined filtering prompt on $765$ videos, 
The \texttt{Gemini-2.0-Flash}'s capability to correctly reject inappropriate content is measured using two complementary metrics: binary rejection accuracy and reason alignment.
Binary rejection evaluates whether \texttt{Gemini-2.0-Flash} and the human annotator agree on the overall accept/reject decision for a video. 
It answers the question: \emph{``Did the model and the human make the same binary decision?''}.
The model achieves a binary evaluation accuracy of 84.3\%, with 473 true negatives (both human and LLM accepted), indicating strong agreement with human acceptance decisions.
We report \emph{recall} and \emph{precision} in \Cref{tab:binary_reason_metrics}, defined as:
\[
\text{Recall} = \frac{TP}{TP + FN}, \qquad
\text{Precision} = \frac{TP}{TP + FP}
\]
where \(TP\) denotes videos rejected by both the human and the model,
\(FN\) denotes human-rejected but model-accepted videos,
and \(FP\) denotes human-accepted but model-rejected videos.

Additionally, analyzing \textbf{rejection reasons} allows us to assess and answers the question: \emph{``When the model and human both reject a video, how often do they agree on the reason for rejection?''} 
\textbf{Reason Match} counts the number of true positives (both rejected) where the rejection reason was the same, while \textbf{Reason Alignment (\%)} expresses this count as a percentage of all true positives. 
We compute this alignment using straightforward string matching between the model-generated reasons and the human-labeled reasons.
High reason alignment indicates that the model is not only matching human decisions at the binary level, but is also capturing the underlying semantic rationale for those decisions. 
Rejected reason categories for all cases can be seen in \Cref{tab:binary_reason_metrics}, which presents both binary rejection and reason matching results

\section{Knowledge Base} \label{sec:knowledge_base}

\begin{table}[!t]
\centering
\scriptsize
\setlength{\tabcolsep}{4pt}
\begin{tabularx}{\linewidth}{|
>{\centering\arraybackslash}p{0.12\linewidth}|
>{\raggedright\arraybackslash}p{0.10\linewidth}|
>{\raggedright\arraybackslash}X|
>{\raggedright\arraybackslash}p{0.18\linewidth}|
>{\raggedright\arraybackslash}p{0.20\linewidth}|}
\hline
\textbf{Emotion Type} & \textbf{Behavior} & \textbf{Observable Cues} & \textbf{Relevant Activities} & \textbf{Contextual Environment} \\
\hline

\multirow{3}{*}{\parbox{\linewidth}{\centering Playful \\ \textit{\scriptsize(Happy, Relaxed)}}}
& Play Bow & Front legs on ground with rear raised, Tail wagging higher than usual, Panting, Staring & Playing (Engaging in playful interactions) & Another dog/human initiating play \\\cline{2-5}
& Pawing & Ears neutral, Panting, Normal tail wagging, Human-focused attention & Interacting (Seeking interaction or attention from humans) & Human they wish to interact with \\\cline{2-5}
& \multicolumn{4}{c|}{\textbf{\textit{.....}}} \\
\hline

\multirow{3}{*}{\parbox{\linewidth}{\centering Calm \\ \textit{\scriptsize(Relaxed)}}}
& Open Mouth & Squinting, Panting, Ears in normal position, Tongue mostly out & Resting (Calm and relaxed posture) & Calm environment with no significant stimulation \\\cline{2-5}
& Ears at Normal Position & Tail carriage normal, Weight evenly distributed, Possibly lying down & Resting (Calm posture) & Normal setting, calm and quiet environment \\\cline{2-5}
& \multicolumn{4}{c|}{\textbf{\textit{.....}}} \\
\hline

\multirow{3}{*}{\parbox{\linewidth}{\centering Curious \\ \textit{\scriptsize(Alert, Inquiring)}}}
& Ears Held Forward & Tail up, Weight on front legs, Attentive eyes, Mouth shut, Ears perked up & Trying to listen & Something intriguing in the environment \\\cline{2-5}
& Sniffing & Head down, Weight on all fours, Tail normal or raised, Soft eyes, Nostrils moving & Curiosity or investigation & Something unfamiliar in the environment, another dog or human \\\cline{2-5}
& \multicolumn{4}{c|}{\textbf{\textit{.....}}} \\
\hline

\multirow{3}{*}{\parbox{\linewidth}{\centering Scared \\ \textit{\scriptsize(Fear, Anxiety)}}}
& Pacing & Ears held back, Drooling, Squinting, Long howls, Moving side to side, Panting & Anxious or restless movement & Being left alone at home or in an unfamiliar space \\\cline{2-5}
& Prey Bow & Tail up, Weight on front paws, Panting, Whale eyes, Piloerection, Snarling & Defensive posture due to anxiety or fear & Something concerning in the environment, accompanied by growling \\\cline{2-5}
& \multicolumn{4}{c|}{\textbf{\textit{.....}}} \\
\hline

\multirow{3}{*}{\parbox{\linewidth}{\centering Aggressive \\ \textit{\scriptsize(Anger, Danger)}}}
& Tail Wagging & Ears forward, Front-weighted posture, Piloerection, Growling, Tail low or between legs, Wagging side to side & Aggression triggered by repeat stimuli & A repeat trigger that is making the dog aggressive \\\cline{2-5}
& Low Pitch Growly Bark & Wagging tail, Snarling, Squinting, Piloerection, Bark sounds low-pitched and deep & Aggressive vocalization indicating a threat & Another dog, human, or unfamiliar stimulus perceived as a threat \\\cline{2-5}
& \multicolumn{4}{c|}{\textbf{\textit{.....}}} \\
\hline

\multirow{3}{*}{\parbox{\linewidth}{\centering Stressed \\ \textit{\scriptsize(Discomfort, Overwhelmed)}}}
& Excessive Panting & Ears normal or held back, Lying down or standing, Drooling, Shaking, Tongue out & Overheating or stress response & After a walk, play session, or exposure to high temperatures \\\cline{2-5}
& Yawning & Squinting, Panting, Weight shifted to hind legs, Piloerection, Ears held back & Increased yawning due to stress, not tiredness & Something scary or stressful in the environment \\\cline{2-5}
& \multicolumn{4}{c|}{\textbf{\textit{.....}}} \\
\hline

\multirow{3}{*}{\parbox{\linewidth}{\centering Sad \\ \textit{\scriptsize(Boredom, Sadness)}}}
& Lethargy & Drooping ears, Slow movements, Avoidance of interaction, Eyes appear distant & Low energy, lack of interest in activity & Loss of a companion, change in environment, isolation \\\cline{2-5}
& Hiding & Ears back, Tail tucked in, Avoiding human or dog interaction & Seeking secluded spaces & Major environmental change, loss of companion, overstimulation \\\cline{2-5}
& \multicolumn{4}{c|}{\textbf{\textit{.....}}} \\
\hline

\end{tabularx}
\caption{Subset of the expert-curated canine behavior ontology curated by a KPA-certified trainer. Each emotion type groups naturalistic behaviors with their observable visual cues, associated activities, and environmental context.}
\label{tab:dog_behaviors}
\end{table}

\begin{table}[!t]
\centering
\scriptsize
\setlength{\tabcolsep}{9pt}
\renewcommand{\arraystretch}{1.15}
\begin{tabularx}{\linewidth}{@{}|
>{\raggedright\arraybackslash}p{0.16\linewidth}|
>{\raggedright\arraybackslash}X|
>{\raggedright\arraybackslash}p{0.24\linewidth}|
>{\raggedright\arraybackslash}p{0.17\linewidth}|@{}}
\hline
\textbf{Behavior} & \textbf{Observable cues} & \textbf{Context} & \textbf{\texttt{K9-Bench} task} \\
\hline
\textbf{Play bow}\textcolor{CiteBlue}{\cite{Byosiere2016,Bekoff1995}}
&
Forelegs down/rear up; tail wag; panting {\color{FACSPlum}{(AU25/26; AD19)}}; gaze/stare
&
Play invitation from dog/human; salient stimulus
&
\textcolor{TaskBurgundy}{Interaction Analysis}
\\
\hline
\textbf{Tail wagging}\textcolor{CiteBlue}{\cite{Quaranta2007,Siniscalchi2018}}
&
Ears forward {\color{FACSPlum}{(EAD101)}}; front-weighted posture; piloerection; growl; tail low/tucked
&
Repeated trigger increasing arousal/aggression
&
\textcolor{TaskBurgundy}{Cause--Effect Analysis}
\\
\hline
\textbf{Prey bow}\textcolor{CiteBlue}{\cite{Gahwiler2020,Siniscalchi2018}}
&
Tail lowered/tucked; panting; whale eye {\scriptsize\color{FACSPlum}{(AD1)}}; piloerection; snarl {\scriptsize\color{FACSPlum}{(AU109+110, AU116)}}; lowered body/weight shift
&
Unfamiliar person/dog/noise; threatening or uncertain stimulus, possibly with growling
&
\textcolor{TaskBurgundy}{Context Analysis}
\\
\hline
\textbf{Pacing}\textcolor{CiteBlue}{\cite{Denham2014}}
&
Repeated route; barrier/exit attention; restlessness
&
Confinement; limited exercise or interaction
&
\textcolor{TaskBurgundy}{Action Sequence}
\\
\hline
\end{tabularx}
\vspace{1pt}
\caption{Compact expert-grounded mapping between canine behaviors, DogFACS-aligned observable cues~\cite{Waller2013DogFACS}, behavioral contexts, and \texttt{K9-Bench} task categories. Posture Analysis is motivated directly by the listed visual cues.}
\vspace{-8mm}
\label{tab:ethogram_compact_rebuttal}
\end{table}

In this section, we present the \textit{Knowledge Base} introduced in \Cref{sec:qa_gen} and is detailed in \Cref{fig:knowledge_base}.
We use it for generating the correct answers in \Cref{fig:qa_generation_prompt}.

\begin{figure}[!t]
\centering
\begin{tcolorbox}[colback=violet!3!white, colframe=pink!40!black, title=\texttt{Knowledge Base}, boxrule=0.6mm, sharp corners, enhanced, fontupper=\small]

\textbf{Dog Action Taxonomy}  
\begin{itemize}
    \item feeding, resting/sleeping, playing, walking, exploring, defecation, vocalizing, social interaction  
    \item dog-to-human communication, dog-to-dog communication, human cue response  
    \item object–dog interaction, distress, sexual activities
\end{itemize}
\vspace{0.3em}

\textbf{Object–Action Affordances}  
\begin{itemize}
    \item food bowl / treat dispenser $\rightarrow$ eating, anticipation
    \item dog bed / sofa / carpet / blanket $\rightarrow$ resting, sleeping
    \item toy ball / rope toy / plush toy $\rightarrow$ playing, chewing
    \item door / doorway $\rightarrow$ wants out, defecation intent, alerting
    \item human person $\rightarrow$ social interaction, attention, alerting
    \item another pet $\rightarrow$ play, social, conflict
\end{itemize}
\vspace{0.3em}

\textbf{Spatial Ontology}  
\begin{itemize}
    \item kitchen, bedroom, backyard, park, restricted zone
\end{itemize}
\vspace{0.3em}

\textbf{Posture \& Gait Cues}  
\begin{itemize}
    \item Postures: standing, sitting, lying (sternal/lateral), crouching, play bow, stretching  
    \item Gaits: walking, trotting, pacing, circling, limping, dragging limbs, stiff gait, collapse  
    \item Micro-indicators: tail (high/tucked/rigid), ears (forward/back), head (neutral/low/tilt), hunched back
\end{itemize}
\vspace{0.3em}

\textbf{Vocalization Cues}  
\begin{itemize}
    \item Bark (short/rapid/deep), whine, whimper, growl (steady/playful), howl, yelp, silence
    \item Context rules:  
    \begin{itemize}
        \item door + whining $\rightarrow$ wants out  
        \item play bow + bark $\rightarrow$ play  
        \item growl + stiff posture $\rightarrow$ warning  
        \item silence in a normally vocal dog $\rightarrow$ anomaly  
    \end{itemize}
\end{itemize}
\end{tcolorbox}
\caption{Canine-Centric Knowledge Base for QA Generation Prompting} \label{fig:knowledge_base}
\end{figure}

\section{Question-Answer Pair Generation Pipeline Details}

\renewcommand{\arraystretch}{1.4}
\setlength{\tabcolsep}{8pt}
\begin{table*}[!t]
\centering
\begin{tabular}{>{\raggedright\arraybackslash}p{2cm} p{5cm} p{5cm}}

\hline
{\textbf{Categories}} & \textbf{What is Tested} & \textbf{Question Prototypes}  \\ \hline
\rowcolor{cbf!20}
Posture Analysis & Focuses on recognizing and interpreting canine body posture across video sequences & \textit{Describe the dog’s ear and head position when the stranger enters the park. What does this suggest about its alertness?} 
\\ 
\rowcolor{cbf!20}
 Action Sequence & Focuses on short-horizon temporal
reasoning by decomposing continuous video frames into 
an ordered sequence of actions. & \textit{Trace the steps the dog takes from noticing the toy to engaging in play with the human.} \\ \hline

\hline
\rowcolor{cspu!20}
Context Analysis
& Identifying how the environmental context condition influences the action and body posture. & \textit{How does leash restriction alter the dog’s behavior when an unfamiliar dog enters the park?} \\ 
\rowcolor{cspu!20}
Cause-Effect Analysis & Detecting the trigger events and immediate resulting responses in the subsequent sequences & \textit{What event immediately triggers the dog to nudge its owner repeatedly?} \\ \hline

\rowcolor{csrd!20}
Interaction Analysis  & Predict action/posture changes involving interaction with human or another canine  & \textit{What social behavioral cues suggest that the dog is seeking comfort from the human after the loud noise?} \\ \hline

\end{tabular}
\caption{Proposed canine-centric video question-answer (QA) task categories used in \texttt{K9-Bench} with their respective question prototypes.} \label{tab:task_suites}
\end{table*}

\renewcommand{\arraystretch}{1.0}
\setlength{\tabcolsep}{6pt}
\begin{table*}[!t]
\centering
\begin{tabular}{>{\raggedright\arraybackslash}p{2cm} >{\raggedright\arraybackslash}p{4.5cm} >{\raggedright\arraybackslash}p{6.0cm}} 
\hline
\textbf{Tasks} & \textbf{What is to be focused?} & \textbf{Answer Prototypes} \\ \hline
\rowcolor{cbf!20}
Posture Analysis & Describe body posture, context, meaning, supporting visual or auditory cues, written as continuous naturalistic observation. & \textit{The dog’s ears stand tall and slightly forward with head raised and fixed gaze as the stranger enters the park paired with a pause in movement suggesting alertness and cautious attention} \\ 
\rowcolor{cbf!20}
Action Sequence & Ordered list of atomic steps [minute\_action\_1, ...], each a small observable action. & \textit{[dog turns head toward gate, dog lifts ears, dog rises from sitting, dog trots toward human, tail wags in arcs, dog sniffs shoes]} \\ \hline

\rowcolor{cspu!20}
Context Analysis & Explain how context influences behavior, integrating spatial, object, or social cues. & \textit{With the open door nearby the dog stands with head raised ears alert tail slightly wagging repeatedly looking toward the entrance while staying near the human reflecting curiosity and vigilance} \\ 
\rowcolor{cspu!20}
Cause-Effect Analysis & Describe observed behavior and its immediate trigger with cause-effect reasoning. & \textit{The dog hears the treat bag rustle lifts its head pricks ears forward and trots toward the human signaling anticipation of reward} \\ \hline

\rowcolor{csrd!20}

Interaction Analysis & Describe posture, vocalizations, and cues in social exchanges with humans/dogs. & \textit{When the human calls its name the dog turns its head ears pricked forward tail wagging rapidly and bounds toward the human expressing eager anticipation and desire for engagement} \\ \hline

\end{tabular}
\caption{Answer styles for each task category with representative examples.}
\label{tab:answer_styles}

\end{table*}

In our proposed task suite, we designed total of $7$ prompts covering correct question–answer (QA) generation ($3$ prompts), narration generation ($1$ prompts) and wrong answer generation tasks ($3$ prompts). 
These prompts serve three main purposes: (i) question and correct-answer pair generation, (ii) video narration generation,   and (iii) plausible wrong answer generation. 

\definecolor{cbf}{HTML}{7ED957}
\definecolor{cspu}{HTML}{B395FF}
\definecolor{csrd}{HTML}{FF70A6}

\subsection{Question and Correct Answer Pair Generation} \label{appendix:qa_gen}
We employ three distinct prompts for QA generation, where task categories are grouped according to the color-coding scheme in \Cref{tab:task_suites}. 
Categories sharing the same color are processed together in a single API call—for example, \textcolor[HTML]{7ED957}{\textbf{posture analysis and action sequence}}, \textcolor[HTML]{B395FF}{\textbf{context analysis and cause–effect analysis}}, and \textcolor[HTML]{FF70A6}{\textbf{interaction analysis}} as a group.
This grouping ensures coherent and effective QA generation across the task suite with reduced hallucinations.
As shown in \Cref{fig:qa_generation_prompt}, we provide the prototype prompt designed for the first two categories (posture analysis and steps of action). 
For the remaining categories, the prompts are adapted to the intended task by modifying task-specific keywords and incorporating the question types listed in \Cref{tab:task_suites}, along with corresponding answer styles and examples presented in \Cref{tab:answer_styles}. 
This prompt also consists of knowledge base shown in Figure~\ref{fig:knowledge_base}.
All videos are processed along with their audio modality using the \texttt{Gemini-2.5-Flash} model provided in the Gemini API \cite{gemini_video_understanding}.

\begin{figure*}[!t]
    \centering
\begin{tcolorbox}[title=Question and Correct Answer Generation Prompt]  
\begin{lstlisting}[style=smallpromptstyle,basicstyle=\ttfamily\scriptsize]
Instructions for Generating Canine Descriptive Foundations Questions and Answers

# ROLE  
You are an expert Canine Behavioral Analyst specializing in generating advanced examination questions and answers that assess deep observation and reasoning skills.  
Your expertise lies in profiling behaviors, decoding postural cues, and mapping sequential actions and interactions of dogs across extended video observations.  
---

# OBJECTIVE  
- Generate **1 to 8 highly challenging questions and answers** testing **long-term understanding of specific behaviors, postures, and action sequences** across the provided video.  
- Questions must assess the candidates ability to **recall, interpret, and connect behavioral patterns, analyze posture dynamics, and trace the stepwise progression of actions**.  
- Use the three analytical categories (#QUESTION_TYPES). Skip a type only if genuinely not applicable:  
---
#QUESTION_TYPES  
---
#ANSWER_STYLES 
---

# CONTEXT INPUTS  
# Video: You will be provided with a video for analysis.  
# Knowledge Base to Apply: {Knowledge_Base}

---

### PROCEDURE ###
1. **Observation Phase**  
   - Watch the entire video carefully.  
   - Pay attention to dog-to-human, dog-to-dog, and human cue response interactions.  
2. **Behavioral Mapping**  
   - Apply the Dog Behavior Taxonomy and Object Behavior Affordances to classify observed actions.  
   - Note relevant **spatial context**, **posture & gait cues**, and **vocalization cues**.  
3. **Interpretation Phase**  
   - Analyze how these behaviors contribute to **long-term social interactions**, **patterns**, and **relational changes** across the observation.  
   - Focus on interaction sequences and their implications (not isolated single moments).  
4. **Question and Answer Construction**  
   - Generate 1 to 8 challenging reasoning questions across the specified #QUESTION_TYPES.  
   - Ensure each question requires memory recall, synthesis of multiple behavioral cues, and interpretation of **social interaction meaning**. 
   - Generate each correct answer for a respective questions across the specified #ANSWER_STYLES. 
5. **Output Formatting**  
   - Strictly return only the questions and answer in the following JSON-like dictionary list format:  
---

# RESTRICTIONS
- Do NOT ask questions beginning with:
  - "When ... ?"
  - "How many ... ?"
  - "How much ... ?"
- Avoid references to time of day (e.g., "night-time", "morning", "bedtime").

---
### EXAMPLE OUTPUT FORMAT ###
[
 {"question_category":"TYPE-I","question":"","answer":""},
 {"question_category":"TYPE-II","question":"","answer":""}
]

\end{lstlisting}  
\end{tcolorbox}  
    \caption{Question and Correct Answer Generation Prompt Prototype. Additional information in prompt such as the \textit{Question\_Types} accompanied by one or more examples are listed in \Cref{tab:task_suites},  and \textit{Answer\_Styles} accompanied by examples are listed in \Cref{tab:answer_styles}.}
    \label{fig:qa_generation_prompt}
\end{figure*}

\subsection{Video Narration} \label{sec:video_narrations}
\begin{figure*}[!t]
    \centering
    \includegraphics[width=0.9\textwidth, trim=2mm 3mm 2.5mm 2mm, clip]{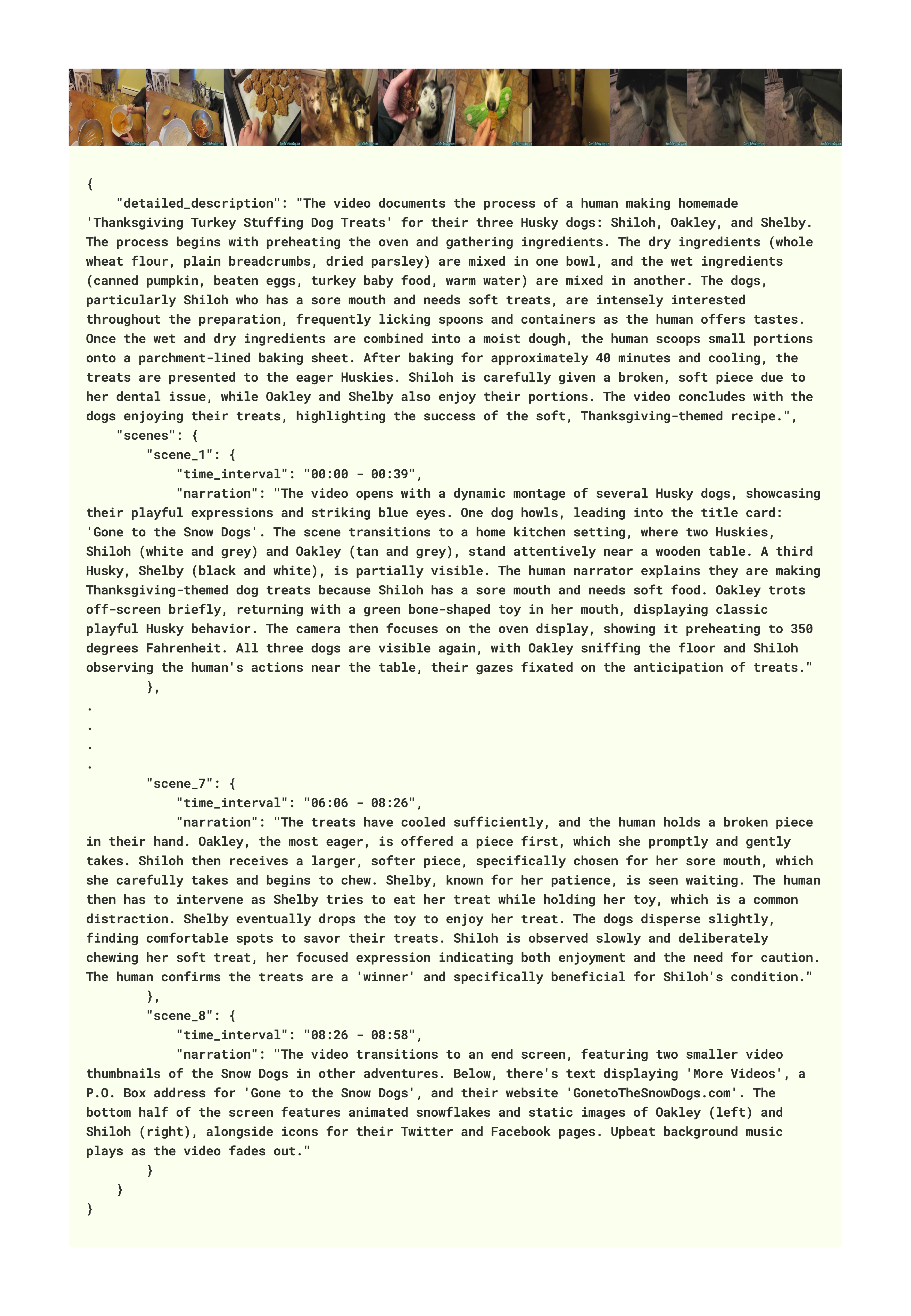}
    \vspace{-3mm}
    \caption{Generated narration for a \texttt{K9-Bench} video, showcasing detailed narration of a video with respect to the canine behavior generated by \texttt{Gemini-2.5-Flash} mention in \Cref{sec:video_narrations}.}
    \label{fig:narration_example}
\end{figure*}

Video Narration is used to generate both high-level video descriptions and fine-grained scene-based summaries.
In the prompt, we define a scene as ``a shift in activity, interaction, or spatial viewpoints."
The prompt for video narration is provided in \Cref{fig:video_narration}. 
All videos are processed along with their audio modality using the \texttt{Gemini-2.5-Flash} model provided in the Gemini API.
The example of the video narration generated on a video is shown in \Cref{fig:narration_example}.

\begin{figure*}[!t]
    \centering
\begin{tcolorbox}[title=Video Narration Generation]  
\begin{lstlisting}[style=promptstyle,basicstyle=\ttfamily\scriptsize]
You are an expert video annotator and canine behavior analyst. Your task is to produce a dense, veterinary-grade narration of the given long video. The narration must integrate environmental context, subject details, actions, and subtle behavioral cues into a coherent account, ensuring minute observation of every scene.

Instructions:
1. Holistic Review 
   - Watch the entire video carefully to understand the complete flow.
   - Write a "detailed_description" that captures the full storyline in a rich, continuous manner.

2. **Scene Segmentation**  
   - Divide the video into scenes with clear time intervals (mm:ss - mm:ss).
   - A scene is defined as a shift in activity, interaction, or spatial arrangement.
   - Each scene should contain minute details, including micro-behaviors and subtle transitions, not just major actions.
   
3. **Scene Narration**  
   - For each scene, provide a dense narration that integrates:
   	- Spatial Context (environment, setting, layout, background changes).
   	- Subject Description (dog's breed, size, coat, markings, humans/other animals).
   	- Action & Behavior (postures, gait, micro-movements, ear/tail/head orientation, anticipatory actions or cues, gaze shifts, vocalizations, stress/displacement cues, affiliative/avoidant tendencies, interactions).
   - Narration must be continuous prose, not bullet points, and reflect fine-grained behavioral tracking.
   - The time interval is metadata only and should not be repeated in the narration text.

#Output Format:
{
  "detailed_description": "Comprehensive storyline of the entire video with dense details.",
  "scenes": {
    "scene_1": {
      "time_interval": "00:00 - 00:42",
      "narration": "Dense narration covering minute details of the scene, integrating spatial context, subject description, and action/behavior."
    },
    "scene_2": {
      "time_interval": "00:43 - 01:27",
      "narration": "Dense narration with fine-grained behavioral details and micro-level transitions."
    }
  }
}
\end{lstlisting}  
\end{tcolorbox}  
    \caption{Video Narration Generation Prompt. See details in \Cref{sec:video_narrations}.}
    \label{fig:video_narration}
\end{figure*}

\subsection{Plausible Wrong Answer Generation} \label{appendix:plausible_gen}
Similar to the correct QA generation process in \Cref{appendix:qa_gen}, three prompts were also developed and utilized here, corresponding to the three task sets based on color coding in \Cref{tab:task_suites}.
Using \texttt{Gemini-2.5-Flash} the four plausible wrong answer are generated. 
The prototype prompt for plausible wrong answer generation is presented in \Cref{fig:wrong_answer_gen_prompt}. 
Similarly, it can be extended to the other two categories by modifying the necessary elements according to the intended task. 
Plausible wrong answer generation prompt utilizes the video narration generated from \Cref{fig:video_narration} and the QA list from \Cref{fig:qa_generation_prompt} as inputs (\Cref{fig:qa_generation} of main text), so we dont supply raw video frames in the VLM context in this step.
Based on these inputs, the prompt generates four plausible wrong answers for each question, thereby producing MCQs with five options, including one correct answer. 

After this step, in total $8263$ QA pairs were generated across $923$ videos.

\begin{figure*}
    \centering
\begin{tcolorbox}[title=Plausible Wrong Answers for MCQ Generation Prompt]  
\begin{lstlisting}[style=smallpromptstyle,basicstyle=\ttfamily\scriptsize]
Instructions for Generating Plausible Wrong Answers for Canine Behavior Multiple Choice Questions
#ROLE 
You are an expert Canine Behavioral Analyst tasked with generating plausible but incorrect answers for pre-existing multiple choice questions (MCQs) designed for a college-level course on canine behavior. 
Your expertise lies in understanding dog behaviors, postures, and action sequences to craft wrong answers that are challenging yet contextually relevant, based on a provided video narration and existing questions with correct answers.
---
### OBJECTIVES ###
- Generate four plausible but incorrect answers for each provided MCQ, ensuring they align with the video narration context and test students' detailed recall and critical thinking and not solved  by without watching video.
- The questions and correct answers are pre-generated, focusing on Behavior Profiling, Posture Analysis, and Steps of Actions as defined in the QA generation prompt.
- Wrong answers should be plausible, varied, closely resemble the correct answer, yet be incorrect, without hinting at the correct choice, and must follow canine behavior, posture, and action sequences.
---
#QUESTION_TYPES:  
---
#ANSWER_STYLES:  
---
### STEPS ### 
1. **Review Video Narration and Questions**
    ......
2. **Correct Answer Protocol**	
    ......
3. **Craft Four Plausible Wrong Answers**
   - For each MCQ, create four wrong answers that are:
   	- Linked Interpretive and Behavioral Alignment: ..........
   	- High Plausibility: ..........
   	- Deceptive Cue Substitution: .........
	- Avoid Blind Model Bias: ..........
   	- Style Consistency: ..........
   	- Length Preserving: ..........
   	- Non-hinting:  ..........
   - Potential Wrong Answer Design by Type: ....... (Instructions specifc to type)
4. **Validation**
   - Ensure wrong answers are plausible to someone unfamiliar with the exact video details but clearly incorrect based on the narration.
    .......
5. **Output Formatting**
    .....
---
### RESTRICTIONS ###
 Do NOT modify the provided question or correct answer.
.......
Do NOT use scene number from the narration.
---
### GENERAL GUIDLINES ###
STRICTLY stay faithful to narrations.
....
STRICTLY Provide the output exactly in the format shown below.
---
### EXAMPLE OUTPUT FORMAT ###
[
  {"question_category": "TYPE-I (unchanged)", "question": "[Provided question text, unchanged]", "correct_answer": "[Provided correct answer, unchanged]", "wrong_answers": ["[Plausible but incorrect answer 1]", "[Plausible but incorrect answer 2]", "[Plausible but incorrect answer 3]", "[Plausible but incorrect answer 4]"]}, ...
]
--- 
### INPUTS ###
## Question and Correct Answer Generated
<List of Question and Correct Answer with Type>
#Video Narration
<Associated Video Narration>
\end{lstlisting}  
\end{tcolorbox}  
\caption{Plausible Wrong Answer Generation Prompt. Video narrations are sourced from the prompt in \Cref{fig:video_narration}, and the corresponding QA list is obtained from \Cref{fig:qa_generation_prompt}. See details in \Cref{appendix:plausible_gen}.}
    \label{fig:wrong_answer_gen_prompt}
\end{figure*}

\section{Bias Mitigation Pipeline}

\subsection{Deaf-Blind LLM Filtering} \label{appendix:deafblind}

The Deaf-Blind LLM performance was evaluated on a total of $9608$ questions using majority voting over three models: \texttt{DeepSeek-R1-Distill-Qwen-32B}, \texttt{Qwen3-32B}, and \texttt{Mistral-Small-3.2-24B-Instruct-2506}. 
This approach ensures high-quality MCQs by preventing models from answering questions solely through prior knowledge or textual shortcuts in QA wording.
The prompt used by these models for deaf-blind filtering is provided in \Cref{fig:deaf_blind_prompt}.
\Cref{tab:deafblind_llm_perf} shows these model performances, which after majority voting reduces the performance to $42.58\%$.

\begin{table}[!t]
\centering
\begin{tabular}{p{5cm}p{2cm}}
\toprule
\textbf{Model} & \textbf{Rejection Rate (\%)} \\
\midrule
DeepSeek-R1-Distill-Qwen-32B & 55.7 \\
Qwen3-32B & 17.05 \\
Mistral-Small-3.2-24B-Instruct-2506 & 52.78 \\
\midrule
\textbf{Majority Vote} & 42.58 \\
\bottomrule
\end{tabular}
\vspace{2mm}
\caption{Performance (rejection rate, \%) of Deaf-Blind LLMs on the dataset.}
\label{tab:deafblind_llm_perf}
\end{table}

\begin{figure*}[!t]
\centering
\begin{tcolorbox}[title=Deaf-Blind LLM Prompt]
\begin{lstlisting}[style=promptstyle,basicstyle=\ttfamily\scriptsize]
You are an expert in answering multiple-choice questions.
You are provided with one question and five answer options (A to E).

Your task:
1. Carefully analyze the questions and options.
2. Provide clear, step-by-step reasoning explaining why each option is correct or incorrect.
4. Select the single best answer (A, B, C, D or E).
Question: {insert question here}
Options:
A. {option A}
B. {option B}
C. {option C}
D. {option D}
E. {option E}

Respond strictly in JSON format as follows:
{
"reasoning": "Detailed step-by-step reasoning comparing all options and showing why the chosen option is correct.",
"answer": "A/B/C/D/E"
}
\end{lstlisting}
\end{tcolorbox}
\caption{Deaf-Blind LLM filtering prompt for multiple-choice question answering without any video inputs.}
\label{fig:deaf_blind_prompt}
\end{figure*}

\begin{figure*}[!t]
\centering
\begin{tcolorbox}[title=Speaker Based Information Removal Prompt,]
\begin{lstlisting}[style=promptstyle,basicstyle=\ttfamily\scriptsize]
You are an expert in refining. Given a question, a set of options (A, B, C, D, E), and the correct answer key (e.g., "B"), convert the options from subjective to objective language. 
- Focus only on directly observable cues such as posture, movement, vocalization, and interaction in the options.
- If the question or options contains speaker or name-specific information (e.g., names of dogs), you may generalize or replace them with neutral terms (e.g., "the dog," "the puppy"), necessary json of their information can be found attached below.
- If the question or options contain any speaker-related information (e.g., "he said," "she mentioned," and more), do not include those references. Instead, reframe the question or statement in a neutral, third-person manner without changing the meaning of correct answer tied to question.
- Remove redundancy across options and make them concise while keeping each one distinct and semantically correct.
- Each option must be a maximum of 20 words.
- Do not change the correct answer key. Keep each option tied to its label.

Provide the response strictly in JSON format:

{{
  "question": "Question (Change only if it contained subject specific information or speaker-related information)",
  "refined_options": {{
    "A": "Refined option A (<=20 words)",
    "B": "Refined option B (<=20 words)",
    "C": "Refined option C (<=20 words)",
    "D": "Refined option D (<=20 words)",
    "E": "Refined option E (<=20 words)"
  }},
  "correct_answer_key": "{correct_answer_key}",
  "validation": "Explain how the refined correct option retains its meaning and remains distinct from the incorrect ones."
}}

# High-Level Information:
{high_level_json}

#Question and Options with Correct Answer:
{qa_json}
\end{lstlisting}
\end{tcolorbox}
\caption{Prompt for removing speaker based information from QA pairs.}
\label{fig:speaker_removal_prompt}
\end{figure*}

\FloatBarrier
\begin{figure*}[!t]
    \centering
    \includegraphics[width=0.9\textwidth, trim=2mm 3mm 2.5mm 2mm, clip]{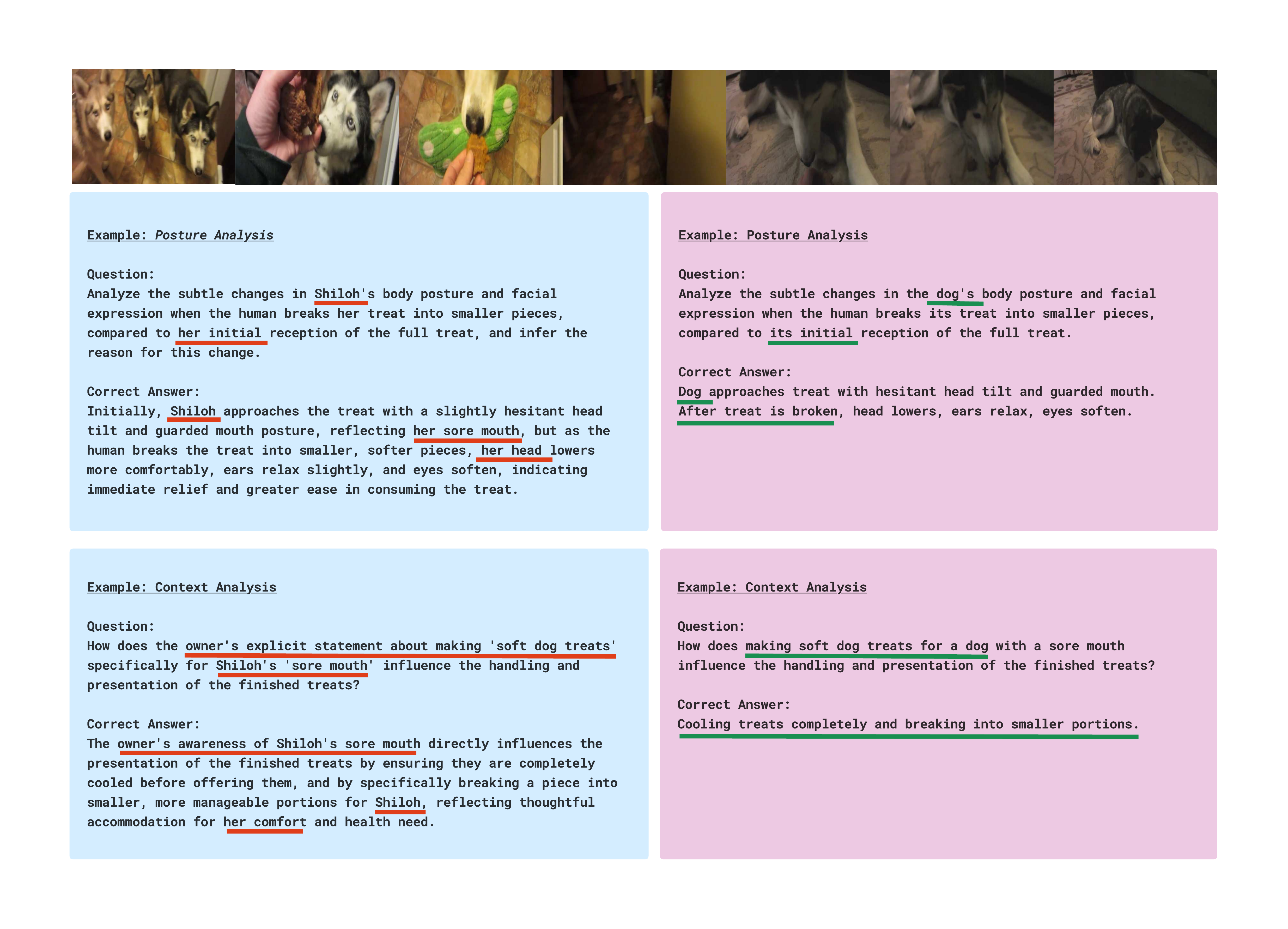}
    \caption{Example of a \texttt{K9-Bench} QA instance showing a before-and-after comparison of bias mitigation. The left shows QA generated by \texttt{Gemini-2.5-Flash} (see \Cref{appendix:qa_gen}), and the right shows the output after applying the bias mitigation pipeline.}
    \label{fig:qa_example}
\end{figure*}
\FloatBarrier

\subsection{Speaker-Based Information Removal} \label{appendix:speaker_information_removal}

Because the QA generation process initially includes the audio modality—introducing speaker information, dog names, and other auditory cues into the QA wording (see \Cref{fig:qa_example})—and relies extensively on the \texttt{Gemini} family of models, we take steps to ensure that the final QA pairs can be fairly interpreted by vision-only or vision–language models. 
To avoid bias arising from \texttt{Gemini} and access to audio, all references to speakers and speaker based information are systematically removed.
This process also aimed to reduce verbosity, enhance clarity, and improve human interpretability without altering the semantic intent of the original QA content. 
The refinement was conducted using \texttt{GLM-4.5-Flash} \cite{zeng2025glm}, which was applied to the entire dataset obtained after deaf-blind filtering. 
The prompt used for removing speaker-related information is shown in \Cref{fig:speaker_removal_prompt}. 
In this prompt, the questions and answer options are provided for refinement, along with high-level contextual information—such as the dogs’ names, coat colors, and breeds—structured as a JSON object. 
This metadata is extracted from the video narration generated using \texttt{Gemini-2.5-Flash} (\Cref{fig:video_narration}).
Example outputs of applying the bias mitigation step can be seen in \Cref{fig:qa_example}.

\section{Benchmark Evaluation Prompts and Configuration}
\label{app:eval_details}

\begin{figure*}[!t]
    \centering
\begin{tcolorbox}[title=MLLM Evaluation Prompt]
\begin{lstlisting}[style=promptstyle,basicstyle=\ttfamily\scriptsize]
You are an expert in video understanding and reasoning. Carefully watch and analyze the entire video before answering the question. 

### Instructions:
1. Watch the entire video to identify subtle behaviors, postures, or interactions relevant to the question.
2. Reason step-by-step, explaining how each observation leads to your conclusion.

Return the response strictly in JSON format with two keys:
- `"reasoning"`: A detailed freeform explanation of your logical steps leading to the correct answer.
- `"answer"`: A concise sentence giving the proper answer.

### Question
{question}
\end{lstlisting}
\end{tcolorbox}
\caption{Evaluation Prompt to get free form response from the models. See details in \Cref{app:eval_details}.} \label{fig:mllm_eval_prompt}
\end{figure*}

All models are evaluated starting with the generation of a free-form response.
The free-form responses for the models reported in \Cref{tab:canine-results} of the main text are generated using the prompt shown in \Cref{fig:mllm_eval_prompt}.
The \texttt{Gemini-2.5-Pro} video and audio+video evaluations were conducted with the respective video and audio+video settings, using videos provided at $1$ FPS.
\texttt{Qwen3-VL-235B-A22B-Instruct} was also evaluated at $1$ FPS.
All the models are evaluated with CoT reasoning enabled.
For \texttt{GPT-4o} (\texttt{model: gpt-4o-2024-08-06}) evaluation $32$ frames that are resized to $512$×$512$ were provided.
For all other open-source models evaluated in \Cref{tab:canine-results}, we use a fixed input of $32$ video frames.
For the frame ablation study, models are additionally evaluated with $32$, $64$, and $128$ input frames.

\subsection{Subjective Evaluation} \label{appendix:subjective_eval}
Following VideoEspresso~\cite{han2025videoespresso}, we employ \texttt{GPT-4o} as an LLM-as-a-Judge to evaluate model-generated responses with respect to the ground-truth answers. 
The evaluation is conducted across five dimensions: logical consistency, factual correctness, accuracy, conciseness, and overall response quality, where each dimension is scored on a scale from 1 to 10. 
We report the averaged scores in \Cref{tab:k9bench-llm-as-judge} and \Cref{tab:human-llm-as-judge} to compare the performance of MLLMs and human responses. 
The complete evaluation prompt provided to \texttt{GPT-4o} is shown in \Cref{fig:subjective_eval_prompt}.

\begin{figure*}[!t]
    \centering
\begin{tcolorbox}[title=LLM-as-a-Judge Evaluation Prompt]
\begin{lstlisting}[style=promptstyle,basicstyle=\ttfamily\scriptsize]
You are a scoring assistant for evaluating text quality.

Evaluate the Model Output based strictly on the given Question and Correct Answer.

Question:
{question}

Correct Answer:
{correct_answer_text}

Model Output:
{model_output}

Evaluation Instructions:
Score each of the following aspects on a scale from 1 to 10 (integers only):

1. Logic:
Evaluate how well the reasoning and structure of the response align with the question, and whether the conclusions follow coherently.
- 1-2: Entirely illogical
- 3-4: Inconsistent or poorly structured
- 5-6: Partially logical with minor gaps
- 7-8: Mostly logical with rare issues
- 9-10: Fully logical and coherent

2. Factuality:
Assess the correctness of the information and the absence of factual errors.
- 1-2: Mostly incorrect or misleading
- 3-4: Significant factual inaccuracies
- 5-6: Some minor inaccuracies
- 7-8: Highly factual with rare errors
- 9-10: Entirely factual

3. Accuracy:
Consider how precisely the response addresses the question.
- 1-2: Irrelevant or off-topic
- 3-4: Partially inaccurate
- 5-6: Moderately accurate
- 7-8: Accurate with minimal flaws
- 9-10: Perfectly accurate

4. Conciseness:
Evaluate how effectively the response conveys its message without unnecessary verbosity.
- 1-2: Excessively wordy or incomplete
- 3-4: Moderately verbose or unfocused
- 5-6: Somewhat concise but improvable
- 7-8: Mostly concise with rare verbosity
- 9-10: Perfectly concise and to the point

5. Overall:
Provide an integrated score reflecting the holistic quality of the response.

Output Requirements:
1. First, provide a brief Chain-of-Thought (CoT) explaining the reasoning behind the scores.
   Format exactly as:
   CoT: {{your concise reasoning here}}

2. Then, output ONLY a JSON dictionary in the exact format below:
{{'Logic': X, 'Factuality': X, 'Accuracy': X, 'Conciseness': X, 'Overall': X}}

Do not include any text outside the CoT and the JSON dictionary.
\end{lstlisting}
\end{tcolorbox}
\caption{LLM-as-a-Judge prompt used with GPT-4o for subjective evaluation. The prompt evaluates model outputs across logic, factuality, accuracy, conciseness, and overall quality. See details in Section~\ref{appendix:subjective_eval}.}
\label{fig:subjective_eval_prompt}
\end{figure*}

\begin{figure*}[t!]
    \centering
    \includegraphics[width=0.8\textwidth, trim=30mm 40mm 25mm 40mm, clip]{figures/subjective_eval_analysis.pdf}
    \vspace{-1mm}
    \caption{Qualitative examples illustrating how the LLM-as-a-Judge (GPT-4o) assigns scores along with its reasoning. The figure presents four QA instances, where the top examples correspond to lower scores and the bottom examples correspond to higher scores. Model outputs are highlighted with red and green blocks to indicate incorrect and correct MCQ evaluations, respectively. Ground truth is marked in \textcolor{green}{green} text. Scores above $7$ typically align with correct responses.}
    \label{fig:subjective_eval_analyses}
\end{figure*}

\begin{figure*}[t!]
    \centering
    \includegraphics[width=0.8\textwidth]{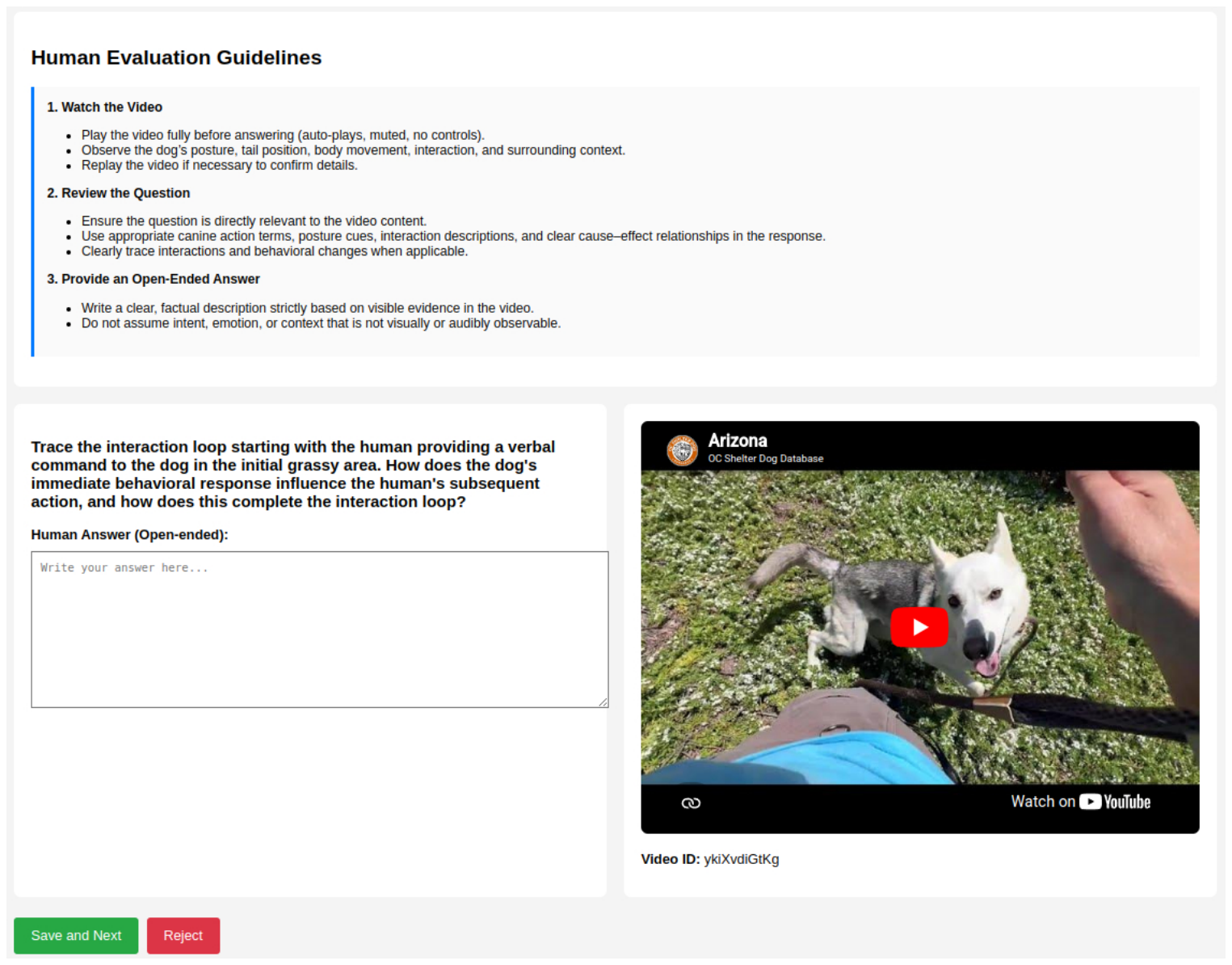}
    \vspace{-1mm}
    \caption{GUI provided with instruction to human annotator to provide free form response.
    }
    \label{fig:human_gui}
\end{figure*}

\section{Computational Setup} \label{sec:computational_setup}

For all open-source evaluations and deafblind evaluations, we utilise a high-performance computing setup consisting of 128 CPU cores and 8 NVIDIA A40 (48\,GB) GPUs. The system is equipped with 512\,GB of SSD storage to support efficient data loading and preprocessing during experimentation. All experiments are implemented using the PyTorch framework with mixed-precision training in FP16 to accelerate computation and reduce memory overhead.

\section{Qualitative Examples of Multimodal Understanding Failures}
\label{appendix:qualitative_analyses}

To quantify and identify the bottlenecks of the best-performing MLLM, we analyzed errors on 105 VQAs and categorized them into four distinct types
This analysis was performed using responses from \texttt{Gemini-2.5-Pro (audio + visual)}, the model that achieved the highest overall accuracy on our benchmark.

The error categories are defined as follows:
\begin{enumerate}
    \item \textbf{Observable Cue Misinterpretation} – misreading or overlooking the dog’s posture, ears, tail, eyes, head, body orientation, or spatial cues (see \Cref{fig:qualitative_observable_cue_misinterpretation} for an example involving missed vocal cues and turning behavior).
    
    \item \textbf{Action Sequence Error} – incorrectly identifying the chronological order of events or actions (see \Cref{fig:qualitative_action_sequence_error} for an example involving incomplete sequence understanding).
    
    \item \textbf{Causal Misattribution} – errors in determining the cause of an action, including misattributing triggers and effects (see \Cref{fig:qualitative_causal_misattribution} for an example involving incorrect causal reasoning).
    
    \item \textbf{Overgeneralization Errors} – providing vague or overly broad descriptions instead of precise, context-specific details (see \Cref{fig:qualitative_overgeneralization_error} for an example involving irrelevant or imprecise focus).
\end{enumerate}
Summary statistics for these 105 VQAs are provided in \Cref{fig:failure_analysis}.

\begin{figure*}[t!]
    \centering
    \includegraphics[width=0.8\textwidth, trim=30mm 40mm 25mm 40mm, clip]{figures/fig_qa_a.pdf}
    \vspace{-1mm}
    \caption{Failure examples illustrating observable cue misinterpretation. See details in Section~\ref{appendix:qualitative_analyses}. Model outputs are highlighted with red and green blocks to indicate incorrect and correct MCQ evaluations, respectively. Ground truth is marked in \textcolor{green}{green} text.
    }
    \label{fig:qualitative_observable_cue_misinterpretation}
\end{figure*}

\begin{figure*}[t!]
    \centering
    \includegraphics[width=0.8\textwidth, trim=30mm 40mm 25mm 40mm, clip]{figures/Fig_qa_b.pdf}
    \vspace{-1mm}
    \caption{Failure examples illustrating causal misattribution. See details in Section~\ref{appendix:qualitative_analyses}. Model outputs are highlighted with red and green blocks to indicate incorrect and correct MCQ evaluations, respectively. Ground truth is marked in \textcolor{green}{green} text.
    }
    \label{fig:qualitative_causal_misattribution}
\end{figure*}

\begin{figure*}[t!]
    \centering
    \includegraphics[width=0.8\textwidth, trim=30mm 40mm 25mm 40mm, clip]{figures/fig_qa_c.pdf}
    \vspace{-1mm}
    \caption{Failure examples illustrating overgeneralization error. See details in Section~\ref{appendix:qualitative_analyses}. Model outputs are highlighted with red and green blocks to indicate incorrect and correct MCQ evaluations, respectively. Ground truth is marked in \textcolor{green}{green} text. $^{*}$ denotes that 32 frames are provided as input to those models. 
    }
    \label{fig:qualitative_overgeneralization_error}
\end{figure*}

\begin{figure*}[t!]
    \centering
    \includegraphics[width=0.8\textwidth, trim=30mm 40mm 25mm 40mm, clip]{figures/fig_qa_d.pdf}
    \vspace{-1mm}
    \caption{Failure examples illustrating action sequence error. See details in Section~\ref{appendix:qualitative_analyses}. Model outputs are highlighted with red and green blocks to indicate incorrect and correct MCQ evaluations, respectively. Ground truth is marked in \textcolor{green}{green} text.
    }
    \label{fig:qualitative_action_sequence_error}
\end{figure*}

\begin{figure*}[t!]
    \centering
    \includegraphics[width=0.8\textwidth, trim=30mm 40mm 25mm 40mm, clip]{figures/fig_think_c.pdf}
    \vspace{-1mm}
    \caption{Qualitative example of thinking model illustrating thinking model performed worse than non-thinking models due to thinking capabilities. Model outputs are highlighted with red and green blocks to indicate incorrect and correct MCQ evaluations, respectively. Ground truth is marked in \textcolor{green}{green} text. $^{*}$ denotes that 32 frames are provided as input to those models. 
    }
    \label{fig:qualitative_think_c}
\end{figure*}

\begin{figure*}[t!]
    \centering
    \includegraphics[width=0.8\textwidth, trim=30mm 40mm 25mm 40mm, clip]{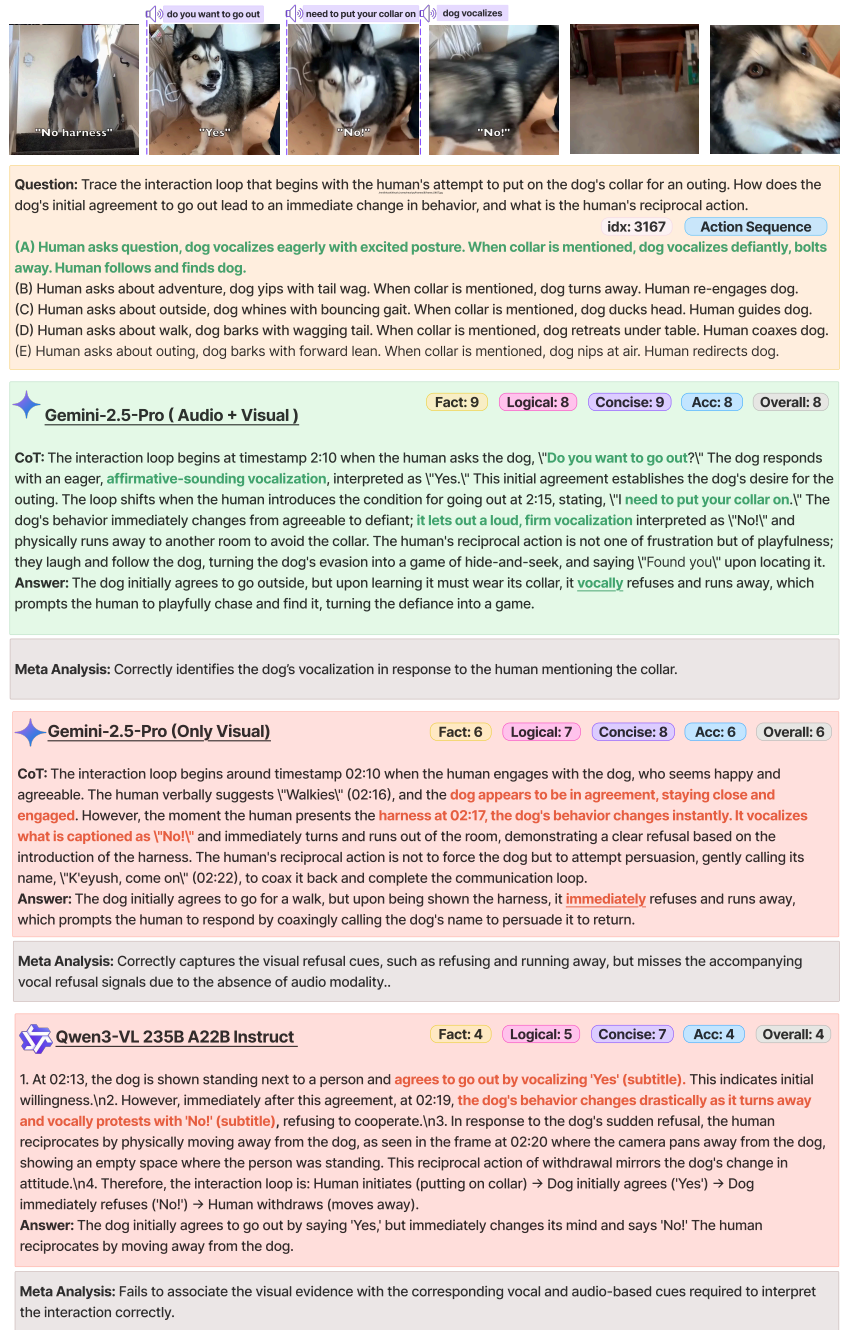}
    \vspace{-1mm}
    \caption{Qualitative comparison showing that both Gemini 2.5 Pro (Audio+Vision) outperform other closed-source VLMs, highlighting the importance of audio modality for accurate interpretation. Model outputs are highlighted with red and green blocks to indicate incorrect and correct MCQ evaluations, respectively. Ground truth is marked in \textcolor{green}{green} text. 
    }
    \label{fig:qualitative_human_b}
\end{figure*}

\begin{figure*}[t!]
    \centering
    \includegraphics[width=0.8\textwidth, trim=30mm 40mm 25mm 40mm, clip]{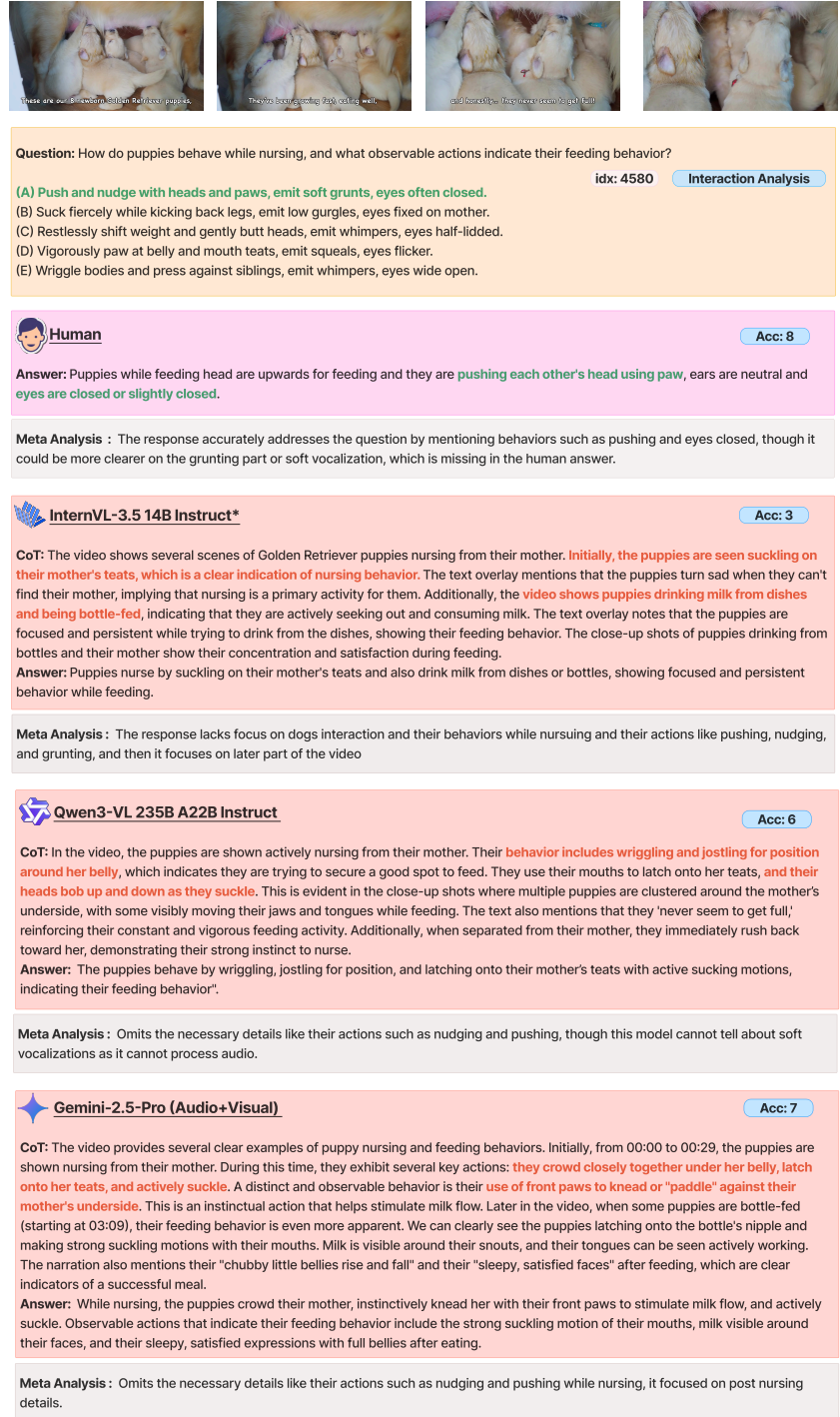}
    \vspace{-1mm}
    \caption{Qualitative comparison of human evaluation and VLM outputs, where human judgment outperforms all evaluated VLMs. Model outputs are highlighted with red and green blocks to indicate incorrect and correct MCQ evaluations, respectively. Ground truth is marked in \textcolor{green}{green} text. $^{*}$ denotes that 32 frames are provided as input to those models.   %
        }
    \label{fig:qualitative_human_a}
\end{figure*}

\section{Limitations} \label{sec:limitations}

The video gathering pipeline may include duplicate videos assigned with different video IDs. 
While this issue is addressed during the human validation stage as one of the rejection reasons, it remains a limitation of the automated pipeline. 
Furthermore, video availability is subject to the public availability of those videos on the hosting platform (YouTube).
\texttt{K9-Bench} tasks and videos do not exhaustively cover all possible canine behaviors and future work should explore more nuanced ways of constructing datasets to capture more fine-grained aspects of canine activity.
In this work, we benchmark zero-shot frontier MLLMs but future research should focus on equipping MLLMs with effective long-horizon temporal reasoning modules through agentic scaffolds or memory-based encoders.

\end{document}